\pdfoutput=1
\documentclass[11pt]{article}

\usepackage[final]{acl}

\usepackage{times}
\usepackage{latexsym}

\usepackage[T1]{fontenc}
\usepackage[utf8]{inputenc}

\usepackage{microtype}

\usepackage{inconsolata}

\usepackage{graphicx}
\usepackage{booktabs}
\usepackage{multirow}
\usepackage{mathtools}
\usepackage{amsmath}
\usepackage{amssymb}
\usepackage{tabularx}
\usepackage{graphicx}
\usepackage{subcaption}

\newcommand{\modelName}{\textsc{MAFIG}}

\title{A Multi-Agent Framework for Feature-Constrained Difficulty Control\\in Reading Comprehension Item Generation}

\author{
    Seonjeong Hwang$^1$,
    Jun Seo$^1$,
    Hyounghun Kim$^{1,2}$, 
    Gary Geunbae Lee$^{1,2}$ \\
    $^1$Graduate School of Artificial Intelligence, POSTECH, Republic of Korea\\
    $^2$Department of Computer Science and Engineering, POSTECH, Republic of Korea\\
    \texttt{\{seonjeongh, sjin4861,
    h.kim, 
    gblee\}@postech.ac.kr} \\
}

\begin{document}
\maketitle
\begin{abstract}

Recent studies in difficulty-controlled reading comprehension item generation have leveraged large language models (LLMs) to produce items by adjusting difficulty-related features.
However, existing methods typically rely on a single-agent prompting approach, which often fails to consistently satisfy specified feature constraints, resulting in items that deviate from the target difficulty level.
To address this limitation, we introduce \modelName, a Multi-agent Framework for Feature-constrained Item Generation, where multiple LLM agents and feature-specific evaluators collaborate to generate and iteratively revise items based on intended constraints.
Furthermore, to verify the efficacy of \modelName\ in difficulty control, we propose a method for constructing a sequence of feature constraint sets that yield items with monotonically increasing difficulty.
Experimental results demonstrate that \modelName\ generates items that adhere to target constraints at a significantly higher rate than baselines, achieving robust difficulty control through the difficulty-calibrated constraint sequence.

\end{abstract}

\section{Introduction}

\begin{figure}[t]
\centering
\includegraphics[width=\columnwidth]{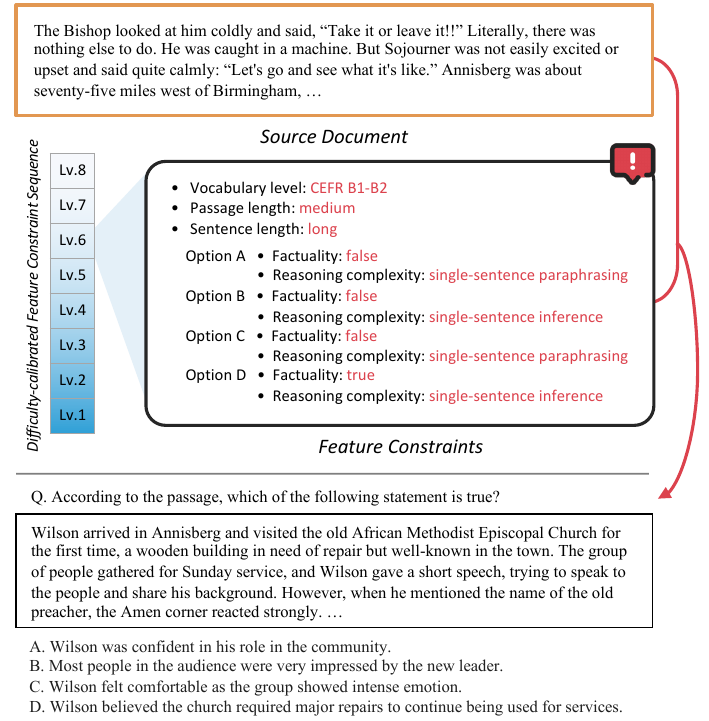}
\caption{\label{fig:framework} Example of feature-constrained difficulty control in multiple-choice RC item generation.}
\label{fig:example}
\end{figure}

Reading comprehension (RC) items are crucial in both language education and proficiency assessment. 
With the continuing expansion of e-learning and computer-based testing, there is a substantial need for methods that can automatically generate high-quality items encompassing a broad spectrum of difficulty levels. 
Recent studies have confirmed that large language models (LLMs) can generate linguistically fluent and pedagogically sound RC items~\cite{xiao2023evaluating,bezirhan2023automated,lee2024few,mucciaccia2025automatic}. 
Nevertheless, the fine-grained control over item difficulty using LLMs remains largely underexplored.

Prior research on difficulty control for RC item generation primarily follows two paradigms.
The first approach utilizes statistical frameworks, such as item response theory (IRT)~\cite{lord1980irt}, to assign difficulty parameters and subsequently train difficulty-aware generative models~\cite{uto2023difficulty,tomikawa2024difficulty,tomikawaB2024difficulty}.
Although this approach enables psychometrically calibrated control, it necessitates substantial learner response data and suffers from limited scalability across diverse item formats.
In parallel, educational measurement research has long investigated difficulty-related item features through item difficulty modeling to guide human item writers in crafting items with targeted difficulty levels~\cite{ferrara2022response}.
Building on this, the second paradigm involves manipulating these features—such as Bloom’s cognitive levels \cite{bloom1956taxonomy} or linguistic attributes like word count and vocabulary level—to modulate difficulty~\cite{elkins2023useful,hwang2024towards,yaacoub2025assessing,chen2025kaqg,oka2025systematic}.
While the robust instruction-following capabilities of LLMs offer promise in this direction, existing methods largely rely on direct prompting or stochastic sampling.
Consequently, they often fail to adhere to specified feature constraints, thereby undermining the reliability of difficulty control.

To bridge this gap, we propose \textbf{\modelName}, a \textbf{M}ulti-\textbf{A}gent framework for \textbf{F}eature-constrained \textbf{I}tem \textbf{G}eneration. 
\modelName\ is designed to generate RC items that strictly conform to multi-dimensional feature specifications as illustrated in Figure~\ref{fig:example}.
The framework operates through a collaborative system of role-specialized LLM agents and feature-specific evaluators.
By leveraging an iterative refinement process, these agents incorporate both external domain knowledge (e.g., standardized vocabulary levels) and their internal reasoning capabilities to ensure rigorous constraint satisfaction.
While \modelName\ is designed for precise adherence to feature constraints, translating this capability into systematic difficulty control necessitates a constraint sequence that predictably yields items across escalating difficulty levels.
Toward this end, we further propose a methodology for constructing difficulty-calibrated constraint sequences, integrating pedagogical principles with empirical verification to ensure a monotonic progression of item complexity.

We evaluate the proposed framework against two baseline approaches: (1) Level-based control, where the LLM generates items based on coarse-grained difficulty indicators (e.g., Level 1 to $N$) relying solely on its internal heuristics; and (2) Feature-based direct prompting, where the LLM is instructed to satisfy all feature constraints within a single-pass generation.
Our experimental results demonstrate that \modelName{} achieves state-of-the-art performance in both constraint satisfaction and difficulty calibration.
Notably, we find that baselines lacking an iterative revision process struggle to satisfy multi-dimensional constraints, leading to inconsistent difficulty alignment—even when leveraging frontier reasoning models such as GPT-5 \cite{openai2025gpt5}.

Our contributions are summarized as follows:
\begin{itemize}
    \item We introduce \modelName, a multi-agent framework that systematically generates RC items that strictly adhere to multi-dimensional feature constraints.
    \item We propose a novel methodology for constructing difficulty-calibrated constraint sequences, enabling the generation of RC items with consistently distinguishable and ordered difficulty levels.
    \item Through extensive experiments, we demonstrate that \modelName\ significantly outperforms baselines in both constraint satisfaction and difficulty calibration. Our results suggest that adherence to fine-grained item features may play an important role in achieving more reliable difficulty control.
\end{itemize}

\section{Related Work}

\paragraph{LLM-based Item Generation and Evaluation.}
Recent advancements in LLMs have facilitated the zero-shot synthesis of test items across diverse pedagogical domains.
Without task-specific fine-tuning, LLMs are capable of producing linguistically coherent and semantically rigorous questions~\cite{elkins2023useful,bezirhan2023automated,lee2024few}.
Beyond generation, contemporary research has explored the role of LLMs as evaluative agents to verify answerability, factual consistency, and distractor quality~\cite{sauberli2024automatic,mucciaccia2025automatic}.
Furthermore, LLMs have been employed as simulated students to analyze item difficulty and pedagogical alignment \cite{lu2024generative,park2024large}.
Collectively, these studies underscore a paradigm shift where LLMs serve as multifaceted components—both as generators and evaluators—within automated assessment pipelines.

\paragraph{Difficulty-Controllable Item Generation.}

Early endeavors in difficulty-controllable generation primarily relied on large-scale datasets labeled with difficulty parameters, often derived from IRT \cite{lord1980irt} or other pedagogical criteria~\cite{gao2018difficulty,uto2023difficulty,tomikawa2024difficulty,tomikawa2024adaptive}.
However, such data-driven methods are costly and frequently lack interpretability regarding the latent factors driving item difficulty.
Consequently, recent research has pivoted toward prompt-based control, where target item types and difficulty levels are specified via natural language instructions.
In particular, prompting LLMs through cognitive taxonomies—such as Bloom’s levels \cite{bloom1956taxonomy}—has been widely explored to align generated items with specific reasoning demands \cite{li2024planning,yaacoub2025assessing}.

Despite their promise, Bloom-level prompting often exhibits inconsistent control over reasoning depth~\cite{elkins2023useful,hwang2024towards}.
Furthermore, given the difficulty variance within identical cognitive levels and the predominance of lower-level cognitive tasks (i.e., Remember and Understand) in high-stakes tests \cite{baghaei2020analysis}, it is evident that such taxonomies are insufficient for achieving fine-grained calibration.
While some recent studies have attempted more granular feature control \cite{chen2025kaqg,oka2025systematic}, they lack a systematic mechanism for refining items when LLMs fail to strictly satisfy specified constraints. Our work bridges this gap by introducing a multi-agent framework that iteratively revises items to ensure rigorous adherence to the feature constraints necessary for precise difficulty control.

\paragraph{Constraint-Satisfaction Generation with LLM Agents.}

Recent advancements have repositioned LLMs as autonomous agents capable of assuming diverse roles. 
By integrating mechanisms such as strategic planning, self-reflection, inter-agent collaboration, and tool-augmented reasoning, these agents can navigate complex tasks and satisfy intricate, user-defined objectives \cite{yao2022react,shinn2023reflexion,madaan2023self,talebirad2023multi}.
Such frameworks have demonstrated substantial efficacy in various constraint-satisfaction tasks, including controllable summarization \cite{ryu2024exploring,retkowski2025zero} and chart generation \cite{li2025metal}.
However, despite these technical strides, the application of multi-agent collaboration to educational assessment—where linguistic, factual, and cognitive constraints must be satisfied simultaneously—remains an underexplored frontier.
Our work bridges this gap by extending the multi-agent constraint-satisfaction paradigm to the domain of RC item generation.
%By doing so, we enable systematic difficulty control through the explicit and iterative alignment of multi-dimensional feature constraints.

\section{Method}

\begin{figure*}[t]
\centering
\includegraphics[width=0.9\textwidth]{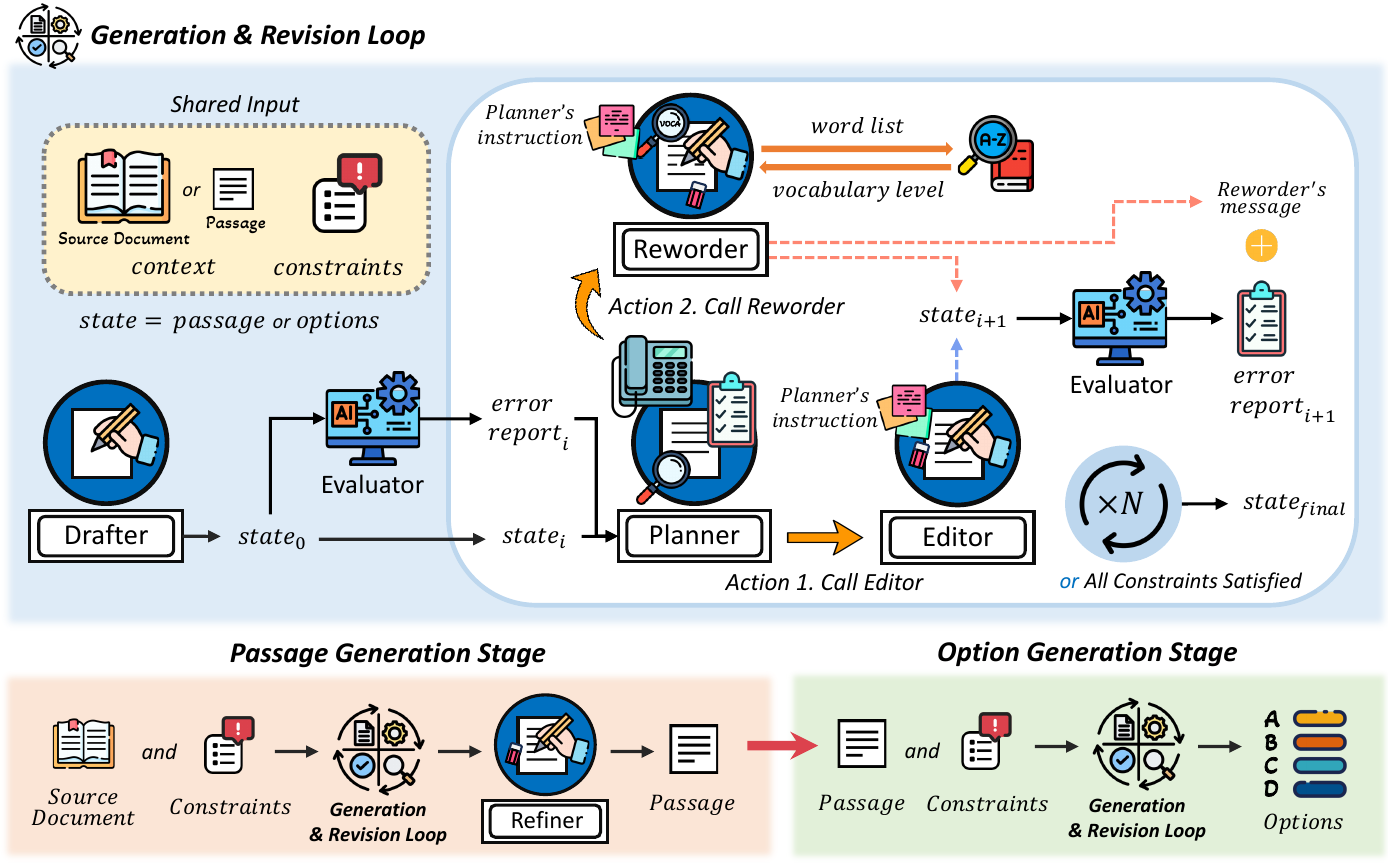}
\caption{\label{fig:framework} Overview of the \modelName\ generation pipeline.
}
\end{figure*}

\subsection{Task Formulation}

In this study, we focus on the multiple-choice factual information (MCFI) format, where test-takers must identify statements that are factually consistent with a given reading passage. 
Specifically, we define an item as a triplet comprising a reading passage, a question stem (e.g., ``According to the passage, which statement is true?''), and a set of options.
Our framework generates an item by taking a source document—which dictates the core content—and a set of feature constraints that determine the target difficulty level as inputs.
Drawing upon established literature that investigates difficulty-related attributes~\cite{bormuth1970children,anderson1972construct,park2004comparison,rafatbakhsh2023predicting}, we formalize six feature variables that govern either cognitive demand or item validity: vocabulary level, passage length, average sentence length, reasoning complexity, factuality, and neutrality. 
Detailed definitions and the operationalization of these features are provided in Appendix \ref{appendix:feature_definitions}.

\subsection{\modelName}

As illustrated in Figure~\ref{fig:framework}, \modelName\ synthesizes multiple-choice RC items through two sequential stages: \textit{Passage Generation} and \textit{Option Generation}.
Each stage incorporates a closed-loop generation and revision mechanism that produces the target component—either the passage or the set of options—while strictly adhering to the specified feature constraints.
In the primary stage, a passage is generated conditioned on the source document and passage-level constraints.
%, such as vocabulary level, passage length, and average sentence length.
This generated passage subsequently serves as the context for the option generation stage.
In this second phase, the framework produces options that satisfy option-level constraints.
%, including neutrality, factuality, and reasoning complexity.
The revision process is performed iteratively until all constraints are met or the predefined maximum iteration threshold is reached.

\paragraph{Evaluator.}

The Evaluator consists of a suite of specialized modules designed to quantify specific feature variables within the generated item. 
This component integrates rule-based modules—leveraging off-the-shelf NLP toolkits—with LLM judges for features requiring semantic understanding.
By measuring the item's attributes, the Evaluator assesses whether the specified constraints are satisfied, and generates a comprehensive \textit{error report} that identifies each violation.
Detailed information is provided in Appendix~\ref{appendix:feature_definitions}.

\paragraph{Drafter.}

The Drafter synthesizes the initial item \textit{state}$_0$ given the source context and targeted feature constraints.
This agent samples multiple independent candidates to facilitate parallel revision, enabling more efficient exploration of the solution space and allowing for early termination.

\paragraph{Planner.}

Conditioned on the current \textit{state}$_i$, the corresponding \textit{error report}$_i$, and a \textit{revision memory} containing plans from previous iterations, the Planner formulates a strategy to revise the item.
This memory mechanism allows the Planner to synthesize revision strategies informed by the history of past modification attempts, thereby avoiding redundant or ineffective edits~\cite{zhang2023large,shinn2023reflexion}.
To further enhance the robustness of this process, we introduce a \textbf{Creativity Enhancement Prompting} strategy.
Specifically, if a particular constraint remains unsatisfied for $t$ consecutive iterations, the Planner is instructed to shift from incremental adjustments to more radical revision strategies—such as excising problematic segments and regenerating them from scratch—thereby breaking the cycle of stagnation.

Following this strategic determination, the planner: (1) selects the optimal agent to invoke—either the Reworder or the Editor—based on the nature of the violation, and (2) generates a granular instruction specifying the necessary modifications.
In the option generation stage, where multiple options are processed concurrently, the Planner further designates which specific option requires intervention and provides tailored instructions.

\paragraph{Reworder.}
The Reworder is specifically tasked with enforcing vocabulary-level constraints, a critical requirement in language proficiency assessments. 
Given that vocabulary regulations often vary across testing organizations and target languages, relying solely on the LLM’s internal heuristics is insufficient for ensuring rigorous alignment. 
To address this, we incorporate a retrieval-augmented generation (RAG) \cite{lewis2020retrieval} that enables the Reworder to revise items by cross-referencing an external, level-specific vocabulary database.

The Reworder takes as input the \textit{context}, the \textit{state}$_i$, the \textit{target vocabulary level}, and the \textit{instruction from the Planner}; during option generation, it additionally receives the \textit{target option}.
The rewording process follows a three-step pipeline:
(1) the Reworder suggests contextually appropriate alternatives for the level-violating words,
(2) a rule-based retriever assigns a vocabulary level to each candidate based on the external database, and
(3) the Reworder replaces the problematic terms with selected alternatives that align with the permitted level range.
In cases where no valid replacements are available, the Reworder notifies the Planner that satisfying the current constraint is infeasible.
Finally, the agent returns the updated $state_{i+1}$, accompanied by a \textit{message} detailing any linguistic bottlenecks encountered during the process.

\paragraph{Editor.}

The Editor revises the item to satisfy all constraints except the vocabulary level.
It takes as input the \textit{context}, \textit{state}$_i$, the set of \textit{constraints}, and the \textit{Planner’s instruction}; during option generation, it additionally receives the \textit{target option}, and finally returns the revised \textit{state}$_{i+1}$.\footnote{
The Editor is unable to accurately assess item features, such as sentence length and reasoning complexity.
In our preliminary experiments, inaccurate self-evaluations from the Editor were found to introduce noise, confusing the Planner during subsequent revisions.
Consequently, the Editor is restricted from sending feedback messages to the Planner.}

\paragraph{Refiner.}
After the iterative revisions, the generated passages are passed through the Refiner.
The Refiner is prompted to make minimal revisions that improve readability and inter-sentence coherence.

\subsection{\label{sec:difficulty_calibrated_feature_constraints} Difficulty-Calibrated Feature Constraints}

While \modelName\ facilitates the synthesis of RC items that adhere to specified feature constraints, achieving fine-grained difficulty control necessitates a difficulty-calibrated feature constraint sequence—a sequence of constraints designed to yield items with monotonically increasing difficulty.
Ideally, such a sequence should be constructed through psychometric analysis.
However, this approach requires an extensive corpus of items with granular feature variations coupled with large-scale learner response data, and such resources are currently not publicly available.
To address this limitation, we propose an alternative methodology that integrates theoretical calibration with empirical verification.

Firstly, we construct initial constraint sets by incrementally adjusting individual feature variables in the direction of increasing cognitive demand.
However, because subtle shifts in cognitive complexity do not guarantee a perceptible change in empirical difficulty, we filter these candidates to ensure that only feature sets yielding consistently distinguishable difficulty are included in the final sequence.
For this purpose, we generate RC items for each candidate feature set using \modelName.
We then perform pairwise difficulty estimation using an LLM judge~\cite{raina2024question} to evaluate the difficulty alignment between items generated under constraint pairs with adjacent theoretical difficulty levels.
%This verification step ensures that the transition between any two successive levels in the sequence corresponds to a statistically significant and human-interpretable increase in difficulty.

A stochastic comparison operator $\mathbb{D}$ takes an ordered item pair $(Q_i, Q_j)$ as input and returns a comparative judgment (i.e., $1$ or $-1$) derived via Chain-of-Thought (CoT)~\cite{wei2022chain} prompting:
\begin{equation}
\mathbb{D}(Q_i, Q_j) = \begin{cases}
1, \text{if } Q_i \succ Q_j \\
-1, \text{if } Q_j \succ Q_i, \\
\end{cases}
\end{equation}
where $\succ$ denotes the ``more difficult than'' relation.
To mitigate positional bias and ensure reliability, we conduct symmetric comparisons across $N$ stochastic inferences.
The Difficulty Alignment Score (DAS) is computed as:
\begin{equation}
\label{eq:da}
\mathrm{DAS}(Q_i, Q_j)=\frac{\sum_{n=1}^N x_f^{(n)} + \sum_{n=1}^N (-x_r^{(n)})}{2N},
\end{equation}
where $x_f^{(n)}= \mathbb{D}^{(n)}(Q_i, Q_j)$ and $x_r^{(n)}=\mathbb{D}^{(n)}(Q_j, Q_i)$ denote the forward and reversed comparison outcomes for the $n$-th stochastic sample, respectively, and ranges from $-1$ to $1$.
We retain only constraint pairs whose score exceeds a predefined threshold $\rho$.
From these validated pairs, we identify the optimal constraint sequence that exhibits a strictly monotonic increase in difficulty.

\section{Experiments}

\subsection{Implementation Details}
We derived an eight-level difficulty-calibrated feature constraint sequence from 16 initial candidate sets by setting $\rho=0.4$ and $N=4$, yielding 8 stochastic inferences in total.
Comprehensive details regarding the calibrated sequence are provided in Appendix~\ref{appendix:detail_calibrated_sequence}.

We employed Qwen3-32B \cite{qwen3technicalreport} in non-reasoning mode to power all LLM agents within \modelName.
The decoding parameters were configured with top-$p=0.8$, top-$k=20$, and a temperature of $0.7$. 
During the initial drafting phase, the number of parallel candidates was set to $5$.
The maximum iteration rounds for passage and option generation were capped at $20$ and $100$, respectively. 
In cases where no candidate achieved full constraint satisfaction within the maximum allowed rounds, the framework returned a randomly selected candidate from the final pool.
Our code and generated items are publicly available at our GitHub repository.\footnote{\textcolor{red}{\url{https://github.com/SeonjeongHwang/mafig}}} Prompt templates used in our experiments are provided in Appendix~\ref{appendix:prompt_templates}.

\subsection{Dataset}
We utilized source documents from the Brown Corpus via the NLTK library.
We randomly sampled 40 documents spanning 10 distinct genres: \textit{news, editorial, reviews, lore, government, fiction, mystery, science fiction, adventure}, and \textit{romance}.
Only the first 50 sentences of each text were used as the source document for item generation.
This selection results in a total of 320 generated items (40 source documents $\times$ 8 difficulty levels).

\subsection{Method Comparison}
Since no existing method generates items at a fine-grained difficulty level in a zero-shot manner, there is no direct baseline for comparison. 
Nevertheless, we construct baselines grounded in the single-pass prompting strategy adopted by most prior work~\cite{elkins2023useful,li2024planning,hwang2024towards,yaacoub2025assessing,chen2025kaqg}, based on two distinct granularities of difficulty control:
\paragraph{Level-based Control.} 
The model is instructed to calibrate difficulty based on an abstract scale.
We utilize two CoT-based prompting strategies: (i) \textbf{Direct Prompting}, where the target level is explicitly specified (e.g., "Generate a level 3 question on a scale of 1–8"), and (ii) \textbf{Incremental Prompting}, which recursively generates a level $i$ item conditioned on the level ${i-1}$ item.\footnote{In incremental prompting, level 1 items serve as initial pivots. Higher-level items are then generated recursively; for fair evaluation, we compare item pairs derived from distinct pivot items.}
\paragraph{Feature-based Control.} 
The model is prompted to satisfy specific feature constraints corresponding to a target difficulty level.
Here, we examine whether rigorous constraint satisfaction enhances calibration robustness by comparing (i) \textbf{Direct Prompting}, which targets predefined features in a single-turn generation, and (ii) \textbf{\modelName}, which employs a multi-agent revision loop until all constraints are met.
We mainly used two different LLMs: Qwen3-32B in non-reasoning mode and GPT-5 with reasoning effort configured to \texttt{medium}.

\subsection{Evaluation Metrics}

We evaluate the performance of each item generation method across three key dimensions: (1) Constraint Satisfaction, (2) Difficulty Calibration, and (3) Item Quality. 
An overview of each metric follows, with formal definitions and mathematical formulations provided in Appendix~\ref{appendix:metrics}.

\paragraph{(1) Constraint Satisfaction.}
This dimension quantifies the extent to which generated items adhere to the specified constraints.
The \textbf{Success Ratio (SR)} measures the proportion of items that satisfy all target constraints simultaneously, while the \textbf{Achievement Ratio (AR)} computes the average fraction of individual constraints successfully met per item.

\paragraph{(2) Difficulty Calibration.}
We assess the model’s ability to control difficulty through the \textbf{Difficulty Alignment Score (DAS)}, which evaluates whether items intended for higher difficulty levels are empirically more challenging, and is derived from both LLM judges and human experts. 
The score ranges from $-1$ to $+1$, where $+1$ indicates perfect monotonic alignment, $0$ signifies inconsistent or negligible differences, and $-1$ represents a complete reversal of the intended difficulty order.
Additionally, we report the \textbf{Complete Alignment Ratio (CAR)}, defined as the proportion of item pairs where a consensus of human experts confirms that the observed difficulty aligns with the intended level.

\paragraph{(3) Item Quality.}
This dimension ensures that the generated items maintain high linguistic and logical standards.
\textbf{Validity} evaluates the generated RC items in terms of their answerability, the correctness of the generated answer against the real answer, and the logical and semantic independence among options.
This is measured via an LLM judge using G-Eval \cite{liu2023g} on a three-point scale (1–3).
Furthermore, we evaluate the \textbf{Coherence} and \textbf{Fluency} of the generated passages using UniEval \cite{zhong2022towards}, with scores normalized between 0 and 1.

%For validation, we prepared an additional set of ten genre-balanced documents, disjoint from the training sources, to assess the consistency of difficulty alignment across unseen topics.  
%This setup allows us to verify whether the proposed framework can robustly handle diverse text types while maintaining feature fidelity and constraint satisfaction.

\section{Results}

\begin{table*}[]
\centering
\small
\begin{tabular}{l|l|cc|c|ccc}
\toprule
\begin{tabular}[c]{@{}l@{}}Difficulty Control\\ Granularity\end{tabular} & Method                           & \begin{tabular}[c]{@{}c@{}}SR\\ (\%)\end{tabular} & \begin{tabular}[c]{@{}c@{}}AR\\ (\%) \end{tabular} & \begin{tabular}[c]{@{}c@{}}DAS\\ {[}-1, 1{]}\end{tabular} & \begin{tabular}[c]{@{}c@{}}Validity\\ {[}1, 3{]}\end{tabular} & \begin{tabular}[c]{@{}c@{}}Coherence\\ {[}0, 1{]}\end{tabular} & \begin{tabular}[c]{@{}c@{}}Fluency\\ {[}0, 1{]}\end{tabular} \\ \midrule
\multirow{4}{*}{Level-based}                                             & Direct$_\mathrm{Qwen3-32B}$      & -                                                                   & -                                                                      & 0.1037                                                                     & 2.6371                                                        & 0.9355                                                         & 0.9280                                                       \\
                                                                         & Direct$_\mathrm{GPT-5}$          & -                                                                   & -                                                                      & 0.2949                                                                     & \textbf{2.9816}                                               & 0.9332                                                         & 0.9408                                                       \\
                                                                         & Incremental$_\mathrm{Qwen3-32B}$ & -                                                                   & -                                                                      & 0.1804                                                                     & 2.5605                                                        & 0.9332                                                         & 0.9408                                                       \\
                                                                         & Incremental$_\mathrm{GPT-5}$     & -                                                                   & -                                                                      & 0.2750                                                                     & 2.9637                                                        & 0.9348                                                         & 0.9309                                                       \\ \midrule
\multirow{3}{*}{Feature-based}                                           & Direct$_\mathrm{Qwen3-32B}$      & \,\,\,0.00                                                                & 59.10                                                                 & 0.2759                                                                     & 2.6094                                                        & 0.9368                                                         & 0.9393                                                       \\
                                                                         & Direct$_\mathrm{GPT-5}$          & \,\,\,2.50                                                                & 77.81                                                                 & 0.4952                                                                     & 2.9105                                                        & 0.9094                                                         & 0.9241                                                       \\
                                                                         & \modelName$_\mathrm{Qwen3-32B}$       & \textbf{92.29}                                                      & \textbf{99.32}                                                        & \textbf{0.5226}                                                            & 2.9242                                                        & \textbf{0.9518}                                                & \textbf{0.9429}        \\ \bottomrule                                     
\end{tabular}
\caption{\label{tab:automatic_evaluation} Automatic evaluation results on diverse difficulty-controllable item generation methods. The best performance of each metric is in \textbf{bold}. Statistics of DAS are reported in Appendix~\ref{appendix:das_std}.}
\end{table*}

\begin{figure*}[t]
\centering
\includegraphics[width=0.8\textwidth]{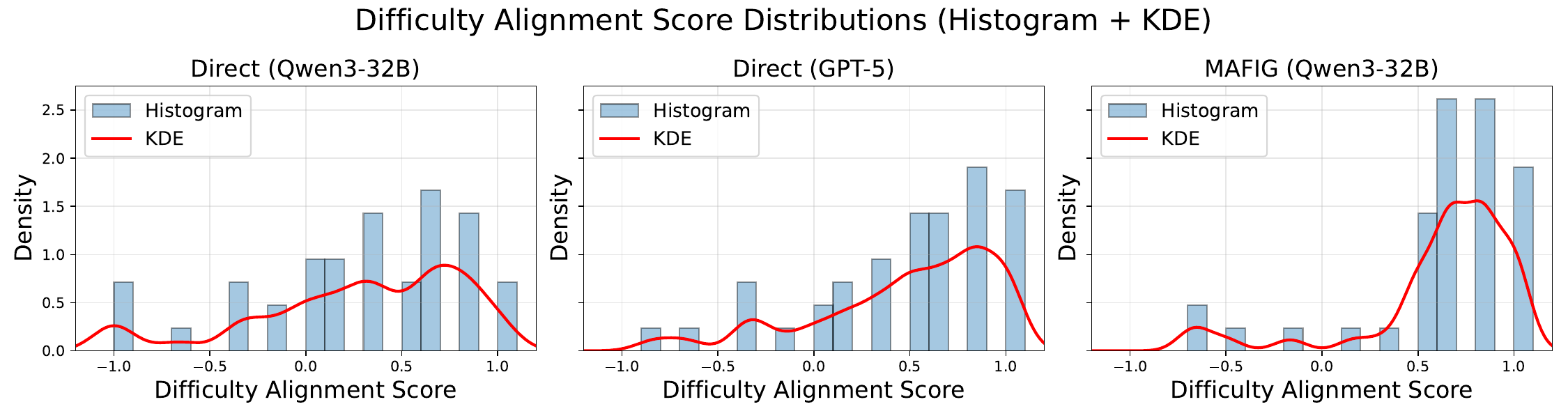}
\caption{\label{fig:human_evaluation_plot} Distribution of Difficulty Alignment Scores assigned by human evaluators.}
\end{figure*}

\subsection{Automatic Evaluation}

\paragraph{\modelName\ effectively generates RC items satisfying feature constraints.}

As summarized in Table~\ref{tab:automatic_evaluation}, \modelName\ consistently outperforms all baseline methods in feature-based difficulty control, achieving a SR of 92.29\%.
This result demonstrates the framework's capability to generate items that strictly adhere to a multi-dimensional constraint set.
In contrast, Direct Prompting with Qwen3-32B fails to produce a single item that satisfies all constraints simultaneously, and even GPT-5 rarely achieves full compliance.
However, an analysis of the AR reveals that Direct Prompting does not entirely disregard instructions; Qwen3-32B and GPT-5 meet over half and three-quarters of the constraints on average, respectively.
These findings imply that while LLMs can partially incorporate feature-based instructions during single-pass generation, they struggle with the simultaneous optimization of multiple constraints.
A case study of constraint satisfaction failures in \modelName\ is provided in Appendix~\ref{appendix:case_study}.

\paragraph{Calibrating RC difficulty remains challenging for LLMs.}

Level-based control methods consistently yield lower DAS compared to their feature-based counterparts.
For Qwen3-32B, both Direct and Incremental Prompting under the level-based setting fail to surpass a score of 0.2. 
In contrast, even the Direct Prompting under feature-based control achieves a higher score of 0.276, despite an AR of only 59.1\%.
A similar trend is observed with GPT-5, which remains below 0.3 in level-based scenarios but reaches 0.495 under feature-based control.
These results underscore that fine-grained difficulty calibration is challenging for even state-of-the-art LLMs when using abstract level descriptions. 
However, explicit feature specification substantially enhances calibration consistency.

\begin{table}[t]
\centering
\small
\begin{tabular}{lcc}
\toprule
Method                      & \begin{tabular}[c]{@{}c@{}}DAS\\ {[}-1, 1{]}\end{tabular} & \begin{tabular}[c]{@{}c@{}}CAR\\ (\%)\end{tabular} \\ \midrule
Direct$_\mathrm{Qwen3-32B}$ & 0.2817                                                                     & 42.86                                                                   \\
Direct$_\mathrm{GPT-5}$     & 0.4722                                                                     & 57.14                                                                   \\
\modelName$_\mathrm{Qwen3-32B}$  & \textbf{0.6190}                                                            & \textbf{76.19}           \\ \bottomrule                                                    
\end{tabular}
\caption{\label{tab:human_evalution_table} Human evaluation results. A statistically significant positive correlation was observed between the LLM judge and the human expert ratings (Spearman’s $\rho = 0.34$, $p < 0.001$) in Difficulty Alignment Score.}
\end{table}

\paragraph{Precise constraint satisfaction drives superior difficulty alignment.}

\modelName$_\mathrm{Qwen3-32B}$ achieves both the highest AR (99.32\%) and the highest DAS (0.5226), whereas Direct$_\mathrm{Qwen3-32B}$ shows the lowest performance in both metrics (59.10\% and 0.2759, respectively). 
Although GPT-5 relies solely on Direct Prompting, it satisfies 77.81\% of the feature constraints on average, resulting in a higher alignment score than Direct$_\mathrm{Qwen3-32B}$.
These results suggest that accurately satisfying difficulty-calibrated feature constraints leads to more reliable difficulty alignment, and that achieving such precise constraint satisfaction requires the iterative revision process of \modelName.

\subsection{Human Evaluation}
We further conducted human evaluation with three domain experts.
We sampled six random source document pairs for each consecutive difficulty level and assessed corresponding RC item pairs generated by three different methods.\footnote{Among the evaluated item pairs, 65.9\% received consistent ratings from all three annotators on the question of which item is more difficult.}
More details are provided in Appendix~\ref{appendix:human_evaluation_details}.

As illustrated in Figure~\ref{fig:human_evaluation_plot}, a large proportion of item pairs generated by \modelName$_\mathrm{Qwen3-32B}$ achieved DAS above $0.5$.
In contrast, while baselines rarely exhibited reversed alignment, many of their pairs fell within the -0.5 to 0.5 range.
This indicates that items at adjacent levels were often indistinguishable in difficulty to human experts.
Furthermore, \modelName$_\mathrm{Qwen3-32B}$ achieved a CAR of 76.19\%, which is substantially higher than that of Direct$_{\mathrm{GPT-5}}$ (57.14\%; see Table~\ref{tab:human_evalution_table}).
Overall, these results demonstrate that the proposed framework can generate item pairs exhibiting perceptible differences in difficulty across fine-grained levels.
They also suggest that when a model fails to satisfy the specified feature constraints, it struggles to establish distinct cognitive demands between adjacent difficulty levels, thereby limiting its calibration accuracy.
Evaluators also identified the difficulty factors driving the differences within each item pair. 
These factors are further analyzed in Appendix ~\ref{appendix:human_justification_analysis}.

\section{Analysis}

\subsection{\label{sec:limitation_of_sampling}Limitations of Direct Prompting with Multiple Sampling}

\begin{figure}[t]
\centering
\includegraphics[width=0.9\columnwidth]{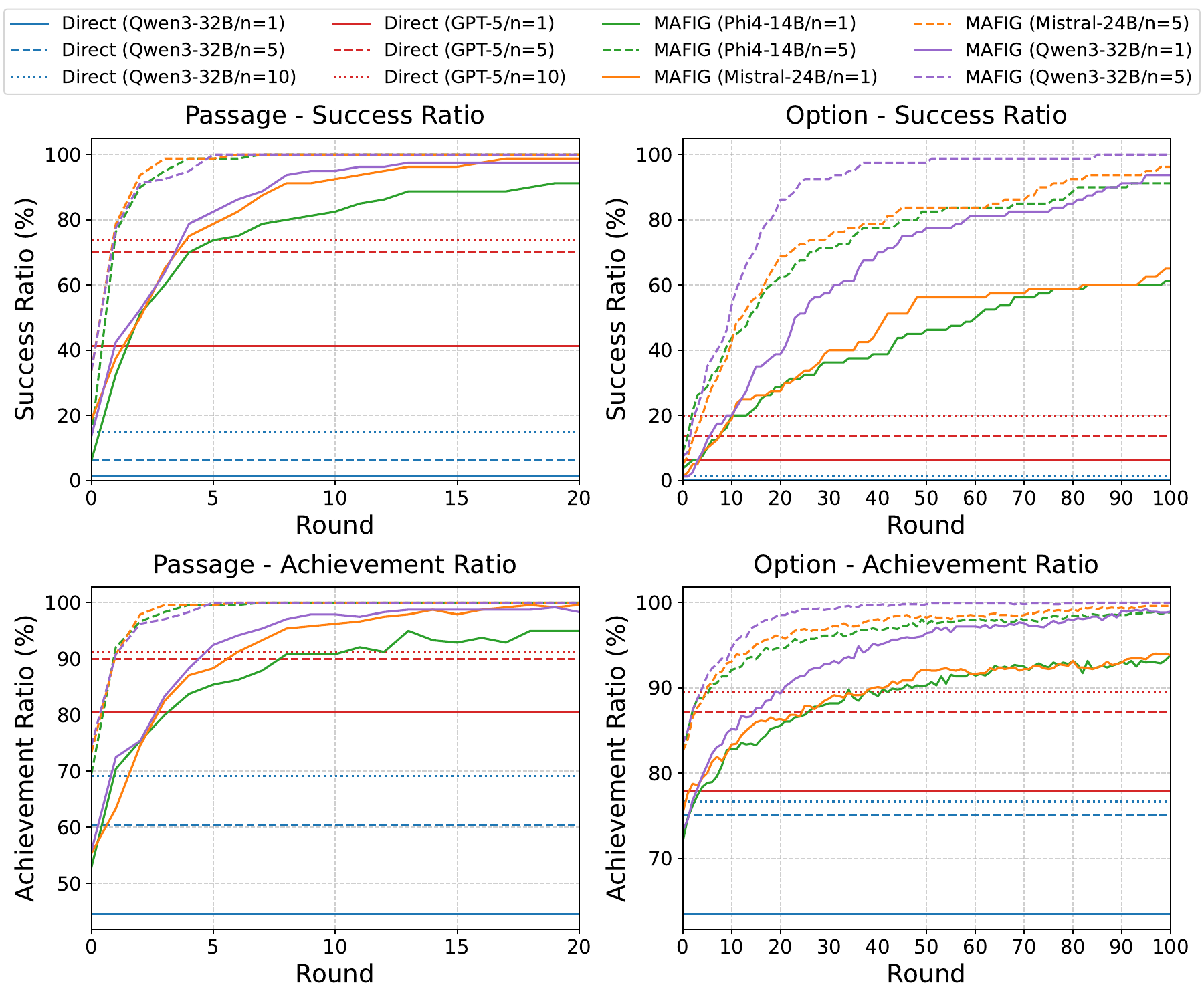}
\caption{\label{fig:method_comparison} Constraint satisfaction performance under different numbers of sampling trials ($n$).}
\end{figure}

We investigated the constraint satisfaction performance of Direct Prompting using multiple sampling. Figure~\ref{fig:method_comparison} illustrates the SR and AR across different numbers of sampling trials (1, 5, and 10) for Qwen3-32B (non-reasoning) and GPT-5 (reasoning). Increasing the number of sampling trials generally improved both SR and AR.
GPT-5 achieved notably high performance, with AR approaching 90\% in both passage and option generation.
However, the performance gain sharply diminished when increasing the sampling times from 5 to 10 compared to the increase from 1 to 5.
This suggests that while stochastic sampling can enhance the likelihood of obtaining partially constraint-compliant items, alone remains insufficient to ensure full constraint satisfaction.
This result underscores the necessity of explicit revision mechanisms.
The feature-wise analysis is covered in Appendix~\ref{appendix:feature-wise}.

\subsection{Model Generalization and Impact of Parallel Revision}

We further evaluated the generalization capability of \modelName\ across diverse backbone LLMs: Qwen3-32B, Mistral-Small-24B~\cite{mistral2025mistral} and Phi-4-14B~\cite{abdin2024phi}.
This experiment also examined the impact of parallel revision, where $n$ item drafts (with $n \in \{1, 5\}$) are sampled and revised in parallel until any single draft fully satisfies the constraint.

As shown in Figure~\ref{fig:method_comparison}, in passage generation, all models except Phi-4-14B achieved a SR of nearly 100\% within 20 rounds, albeit with different convergence speeds.
In contrast, option generation proved more challenging: with only a single draft ($n=1$), all models except Qwen3-32B failed to surpass a 60\% SR, even after 100 revision rounds.
This may be attributed to the reasoning complexity features controlled during option generation, which are more challenging to control than surface-level features.

When the number of initial candidates was increased to $n=5$, all models achieved 100\% success, converging substantially faster—passage generation typically terminated within five rounds.
For option generation, Qwen3-32B required roughly half the number of revision rounds compared to the single-draft setting.
Interestingly, performing parallel revisions on multiple drafts using lighter models (Mistral-small-24B and Phi-4-14B) yielded higher overall constraint satisfaction than revising a single draft using the larger Qwen3-32B model.
This advantage is likely due to the diversity in revision paths across drafts, where different initial drafts require distinct modifications to satisfy the constraints.

\subsection{Ablation Study}

\begin{figure}[t]
\centering
\includegraphics[width=0.95\columnwidth]{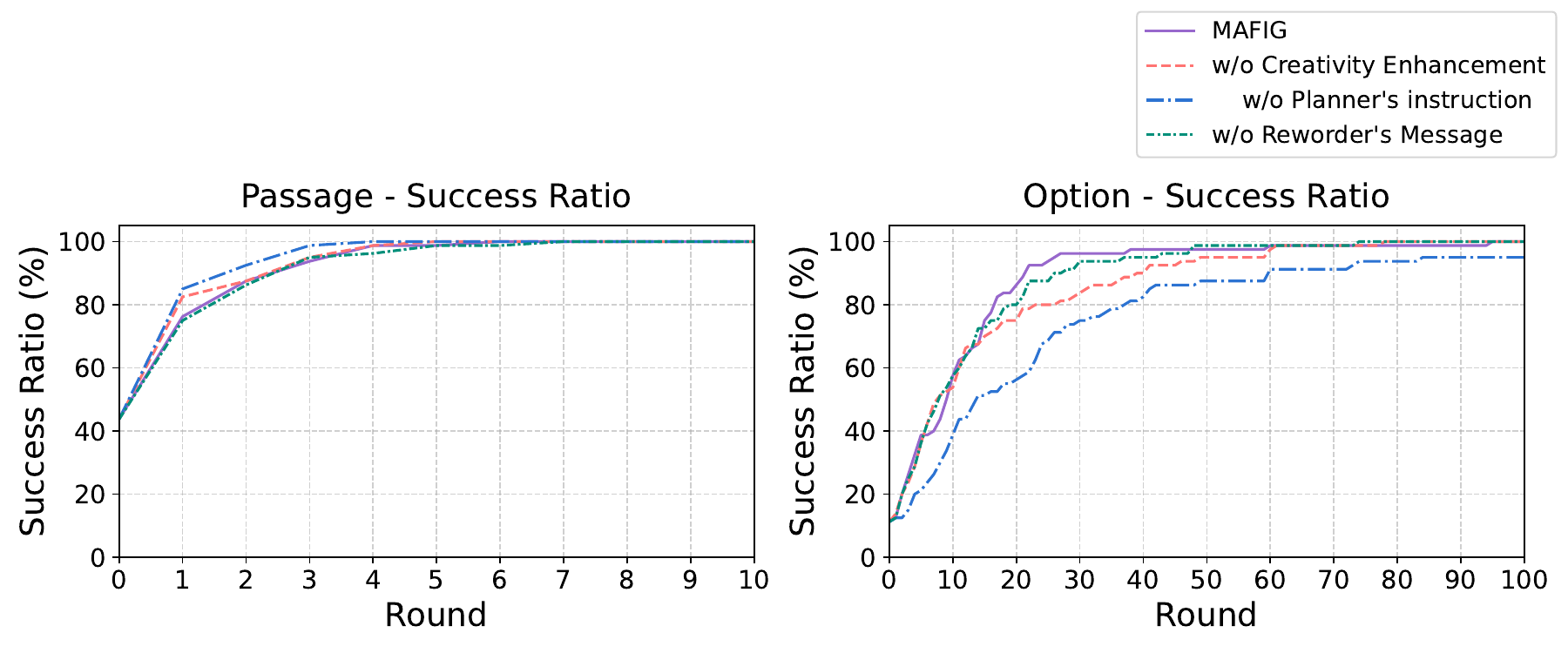}
\caption{\label{fig:ablation} Ablation results showing the effect of Planner’s instruction, Reworder’s message, and Creativity Enhancement on Success Ratio convergence across revision rounds.}
\end{figure}

We conducted an ablation study to assess the contribution of three strategies—Planner’s Instruction, Reworder’s Message, and Creativity Enhancement Prompting—to the iterative revision process in both passage and option generation (Figure~\ref{fig:ablation}).
When the Planner’s instruction mechanism was removed and sub-agents (Reworder and Editor) revised items solely based on error reports, we observed a substantially slower convergence in option generation.
Similarly, disabling the Reworder’s feedback message and the creativity enhancement mechanism also resulted in slower convergence.
In contrast, the passage generation process showed minimal performance differences across ablation settings; in fact, replacing the Planner’s instruction with direct error reports occasionally led to slightly faster convergence.
In summary, while these auxiliary mechanisms did not substantially affect surface-level revisions in passage generation, they proved critical for accelerating convergence in option generation, where more complex semantic and reasoning-based revisions are required to satisfy the constraints.

\section{\label{appendix:computational_cost}Computational Cost}

\begin{figure}[t]
\centering
\includegraphics[width=\columnwidth]{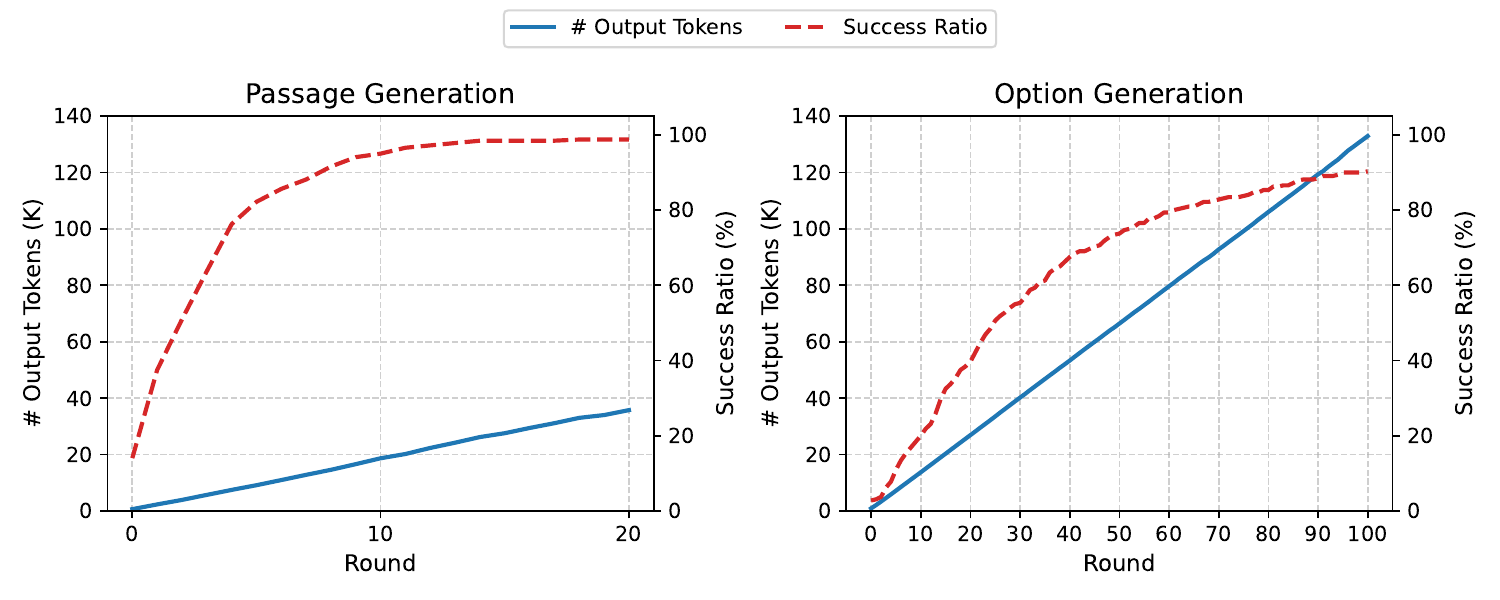}
\caption{\label{fig:output_token_num} Success Ratio and the number of cumulative output tokens across revision rounds in \modelName. %Round 0 represents the initial draft generated by the Drafter.
}
\end{figure}

Unlike baseline methods based on single-pass prompting, \modelName\ employs an iterative revision process to ensure that the generated RC items strictly adhere to multifaceted feature constraints.
While this approach guarantees high-fidelity item generation, it inherently introduces a significant computational overhead compared to non-iterative methods.
Figure~\ref{fig:output_token_num} illustrates this trade-off by comparing the cumulative output tokens against the SR (at $n=1$) for both passage and option generation stages.
In passage generation, the framework demonstrates relatively efficient convergence, with 90\% of cases satisfying all linguistic and structural constraints within 10 rounds at a cost of approximately 20K cumulative tokens.
In contrast, option generation requires a much more intensive search process to satisfy fine-grained distractor constraints; achieving about 90\% SR can necessitate up to 100 rounds, resulting in an accumulated token count exceeding 130K.

These results acknowledge that the substantial overhead and associated latency may render \modelName\ less suitable for real-time applications or large-scale deployments where instantaneous item generation is required.
Nevertheless, the necessity of our iterative framework remains evident because, as discussed in Section~\ref{sec:limitation_of_sampling}, repeated single-pass sampling often fails to converge on a valid solution that satisfies all complex feature constraints simultaneously.
Furthermore, in high-stakes environments such as standardized language assessments or national-level examinations, difficulty calibration and adherence to pedagogical constraints take precedence over cost efficiency.
In such contexts, the increased computational cost can be a justifiable trade-off for the reliability and constraint satisfaction guaranteed by our framework.

\section{Conclusion}
In this study, we introduced \modelName, a multi-agent framework for feature-constrained RC item generation.
\modelName\ coordinates role-specialized LLM agents in conjunction with feature-specific evaluators to iteratively generate, assess, and revise RC items, ensuring strict adherence to all designated feature constraints.
To complement this framework, we proposed a method for constructing a difficulty-calibrated sequence of feature constraint sets, integrating theoretical insights from educational measurement with empirical verification.
Experimental results demonstrated that our framework achieved substantially higher constraint satisfaction rates and exhibited superior difficulty alignment performance, as validated by both LLM-based and human expert evaluations.

%Experimental results demonstrated that even strong reasoning models struggle to fully satisfy multiple feature constraints via a single prompt.

%These findings highlight the importance of explicit feature-level control and iterative generation strategies in building explainable and pedagogically reliable AI-assisted assessment systems.
%Future work will extend this framework to broader question types and adaptive testing scenarios, further exploring the intersection of cognitive theory and controllable text generation.

\section*{Limitations}

\paragraph{Scope of Item Formats and Difficulty Factors.}
Our experiments focus on a single item format—multiple-choice factual information (MCFI) questions—and a fixed set of difficulty-related features.
Item difficulty is influenced by a wide range of factors, and the relative importance of these factors varies across item formats.
We selected MCFI items because they represent a canonical RC format in which both surface-level features and reasoning-related features jointly contribute to cognitive load.
In addition, passage topic and question type were treated as controlled variables to isolate fine-grained difficulty variation within a homogeneous setting.

Consequently, while our findings may not directly generalize to other RC formats, the proposed framework is not inherently restricted to MCFI items. 
Extending the framework to alternative formats would primarily require defining additional feature evaluators and incorporating corresponding descriptions into agent prompts, though the two-stage passage–option generation structure might necessitate adaptation depending on the target format.
Nevertheless, this study is valuable as the pioneering effort to showcase how LLM-based multi-agent frameworks can be leveraged for feature-constrained item generation and subsequent fine-grained difficulty control.

\paragraph{Lack of Psychometric Validation Against Examinee Performance.}
In educational measurement research, a range of difficulty-related item features have been identified and used as practical guidelines for item writers to produce questions at a desired difficulty level~\cite{ferrara2022response}. 
Building on this practice, our framework generates RC items conditioned on feature constraints corresponding to a target difficulty range, and evaluates difficulty alignment through expert judgments — that is, from a \textit{prior difficulty} perspective based on linguistic and cognitive item features, rather than on observed examinee performance. 

Consequently, the generated items are not psychometrically calibrated in an absolute sense: their difficulty has neither been validated against student error rates nor evaluated in terms of item response patterns. 
While the framework is intended as a practical support tool for item writers, future work should incorporate empirical validation through examinee response data to establish psychometric grounding for the difficulty levels produced by our framework.

\paragraph{Computational Cost and Latency.}
The proposed multi-agent framework incurs additional computational cost due to its iterative revision process.
While our experiments show that the number of revision rounds is typically limited, this overhead may still pose challenges in large-scale or time-sensitive deployment scenarios.
As such, the framework is most suitable for settings where reliability and precise difficulty control are prioritized over minimal generation cost.

\paragraph{Reliance on Feature Definitions and Evaluator Quality.}
The effectiveness of \modelName\ depends on the quality of feature definitions and evaluators.
Because difficulty control is achieved through explicit feature constraints, inaccuracies or ambiguities in feature specifications—or erroneous predictions by the evaluators used to measure them—can affect the reliability of constraint satisfaction and difficulty alignment.
In particular, some features, such as reasoning complexity, rely on LLM-based evaluators, which may introduce noise or bias.
This dependency reflects a broader limitation of feature-based item design, even as it offers greater transparency and interpretability in how difficulty is operationalized.

%% measurement in education 연구들에서 item difficulty modeling을 통해 difficulty related item feature들을 연구했고, 이들은 item writer들이 desired range의 difficulty의 문항을 produce하는데 guide를 제공함. 
%% 본 연구는 이러한 현업의 상황을 고려해 desired difficulty에 대응되는 item feature들을 item writer가 구성하면 이에 대응되는 문항을 자동으로 생성할 수 있는 framework를 제공하는 것을 목표로 했으며, difficulty alignment에 대한 평가 또한 전문가의 판단을 기준으로 했음.
%% 다만 실제 학생들의 오답률을 기반으로 생성된 문항들의 난이도를 분석하지는 않았고 여전히 prior difficulty 관점에서 문항들을 구성하는데 그쳤음.
%% 그럼에도 불구하고 본 연구는 item writer들에게 유용한 tool로서 동작할 수 있음, 단순히 quality적으로 타당한 문항을 생성하는데 그치는 것이 아니라 목표한 난이도의 문항을 개발하는 것을 support할 수 있음.

%% 알려진 difficulty related factor는 매우 많으며 Item format마다 적용되는 인자가 조금씩 달라짐. 본 연구에서는 liguistic features부터 reasoning features까지 다양한 난이도 인자의 영향을 받는 대표적인 유형인 MCFI에 집중하여 실험을 수행했음. 본 framework를 더 다양한 문항 유형에 적용하는 것은 간단함. 조절하고자 하는 feature에 대한 evaluator를 추가하고 각 agent의 prompt에 feature에 대한 설명을 추가하면 됨. 다만 본 연구에서는 passage generation과 option generation의 two step으로 구성했으나, 문항 포맷에 따라 구조 수정이 필요할 수 있음

%% 이론 상의 각 feature별 cognitive load를 기준으로 난이도를 매김
%% 실제로는 각 feature마다 가중치를 달리 해서 고려해야할 것이며 현재 이산화된 각 feature별 level range도 더 탐구되어야할것임.
%% 현재 우리는 단순히 각 level의 feature를 구성할 때 feature 하나 혹은 두개의 level을 한단계 씩 높이는 식으로 cognitive load를 높였음. \modelName이 난이도 calibration에 효과적이라는 것을 empirical하게 보이기 위해서.
%% 그리고 LLM Judge를 통해 인지 가능한 수준의 난이도 차이를 보이는 constraint sequence를 구축했는데,
%% 이런식이면 각 level 간의 난이도 gap에 차이가 발생할 수 있고, 실제 교육적으로 필요한 난이도 gap이랑 align되지 않을 수 있음.
%% 우리가 feature를 만족하는 문항을 생성하는 파이프라인을 구축해놓았기 때문에 level별 feature sequence를 찾는 방식을 더 교육학적으로 발전시킨다면 훨씬 교육쪽에서 효과적으로 활용될 수 있을 것임.

\section*{Acknowledgments}

This research was supported by Culture, Sports and Tourism R\&D Program through the Korea Creative Content Agency grant funded by the Ministry of Culture, Sports and Tourism in 2025 (Project Name: Development of an AI-Based Korean Diagnostic System for Efficient Korean Speaking Learning by Foreigners, Project Number: RS-2025-02413038, Contribution Rate: 45\%); by the IITP (Institute of Information \& Coummunications Technology Planning \& Evaluation) - ITRC (Information Technology Research Center) grant funded by the Korea government (Ministry of Science and ICT) (IITP-2026-RS-2024-00437866, Contribution Rate: 45\%); and by Institute of Information \& communications Technology Planning \& Evaluation (IITP) grant funded by the Korea government (MSIT) (No.RS-2019-II191906, Artificial Intelligence Graduate School Program (POSTECH), Contribution Rate: 10\%).

\bibliography{custom}

\newpage

\appendix

\section{\label{appendix:feature_definitions}Feature Definitions and Feature-Specific Evaluators}

We employ six feature variables: four features control the cognitive demand of the item—vocabulary level, passage length, average sentence length, and reasoning complexity—while two ensure validity—factuality and neutrality among options.
Features with continuous values are discretized into categorical ranges: passage length $\in$ \{short (5--10 sentences), medium (11--20 sentences), long (21--30 sentences)\}, average sentence length $\in$ \{short ($<$10 words), medium (10--15 words), long (15--20 words)\}.
Vocabulary level is divided into CEFR-based bands—\{A (A1–A2), B (B1–B2), C (C1-C2)\}.
The reasoning complexity of each option is classified into five levels: \{\textit{single-sentence word matching}, \textit{single-sentence paraphrasing}, \textit{single-sentence inference}, \textit{multi-sentence inference}, \textit{not enough information}\}~\cite{lai2017race,hwang2025can}.
Finally, factuality can take values \{\textit{true}, \textit{false}, \textit{not given}\}, and neutrality must be maintained to ensure that all options are logically independent and collectively valid.

The Evaluators comprise off-the-shelf NLP toolkits and LLM judges.
Passage length and sentence length are computed using  NLTK~\cite{loper2002nltk} library, while vocabulary level is determined by the highest CEFR level among the words contained in a passage or an option.
Following~\citet{hwang2025can}, we evaluate the reasoning complexity of each option along two sub-dimensions, Evidence Scope and Transformation Level, using CoT prompting with self-consistency decoding~\cite{wei2022chain,wang2022self}, achieving Macro F1 scores of 69.0 and 70.8, respectively. 
We use the same strategy to assess the factuality and neutrality of options.
These LLM-based evaluations were conducted using Qwen3-32B~\cite{qwen3technicalreport} in non-reasoning mode.

\section{\label{appendix:metrics}Evaluation Metrics}
\paragraph{Success Ratio.}
The SR measures the proportion of items that satisfy all target constraints simultaneously.
Formally, given a source document $S$ and a \textit{feature constraint set} $C = \{(X_i, x_i)\}_{i=1}^{M}$, where each feature variable $X_i$ is assigned a target value $x_i$,  we generate an item $Q_C$ whose passage is composed based on the content of $S$ while satisfying all feature constraints specified in $C$.
The generation is considered successful if all constraints in $C$ are satisfied as follows:
\begin{equation}
\label{eq:success_rate}
\begin{aligned}
& \text{Success}(Q_C) \\
& = \mathbb{I}\big[\, \forall i \in \{1, \ldots, M\},\, E_{X_i}(Q_C) = x_i \,\big],
\end{aligned}
\end{equation}
where $E_{X_i}(\cdot)$ denotes the evaluation function that measures the realized value of the feature variable $X_i$ for the generated item.

In our experiments, an item generation task is considered successful if at least one of the $n$ parallel drafts satisfies all target constraints. Constraint satisfaction was verified using the same automated evaluators integrated within \modelName.

\paragraph{Achievement Ratio.}  
To capture partial success, we measured the proportion of feature constraints satisfied by the generated item.  
Formally, this is calculated as the percentage of satisfied constraints in set $C$:
\begin{align}
\frac{\sum_{i=1}^M \mathbb{I}[E_{X_i}(Q_C) = x_i]}{M}\times{100}
\end{align}
This metric reflects how closely a generated item aligns with the target specifications even when full satisfaction is not achieved.
When multiple parallel candidate items are generated ($n > 1$), we report the results based on the item achieving the maximum AR.

\paragraph{Difficulty Alignment Score.}
The DAS measures whether the given item pairs, which has the adjacent levels, have the difficulty order corresponding to the intended order.
We report the average DAS that are measured for the item pairs sampled from the adjacent difficulty levels (level-$i$ and level-$i+1$) from the identical source documents.

For evaluation using \textbf{LLM judges}, we adopted the pairwise comparison method defined in Equation~\ref{eq:da}. 
GPT-5-mini~\cite{openai2025gpt5} served as the difficulty evaluator, with sampling parameters set to $N=4$, temperature $1.0$, and top-$p$ $1.0$.
A score near $1$ indicates that the method can perfectly calibrate item difficulty. Conversely, a score near $-1$ implies that the method consistently controls difficulty in reverse of the intended order; that is, the item generated for Level $i$ is perceived as more difficult than the one generated for Level $i+1$. 
A score of $0$ signifies that the LLM judge exhibits low confidence, resulting in inconsistent outputs.

In the evaluation with \textbf{human experts}, we similarly employed a pairwise estimation framework where annotators identified the more difficult question within a pair.
If the item intended to be harder was correctly identified, the pair was assigned a score of $+1$; otherwise, it received $-1$.
To evaluate whether the difficulty gap was non-trivial, annotators assigned one of three labels: (1) \textit{Almost no difference}, where the items are nearly identical in difficulty; (2) \textit{Moderate difference}, representing a gap that distinguishes ``easy'' and ``hard'' items for students of the same proficiency level; and (3) \textit{Large difference}, indicating a substantial gap suitable for students of different proficiency levels.

Unlike statistically derived parameters, human perception of difficulty is inherently subjective and an individual's judgment of a ``difficulty gap'' can vary; some may perceive items within the same proficiency range as identical, while others may discern subtle nuances.
We intentionally employed this three-level scheme to capture such fine-grained distinctions, encouraging annotators to recognize and report even minor variations. 
This granular approach ensured that evaluators remained sensitive to subtle differences during the pairwise comparison. 
For the final alignment metric, however, we consolidated these into two categories to focus on the presence of an educationally significant difference: \textbf{Case 1} (Almost no difference) and \textbf{Case 2} (Moderate or Large difference).
This binary distinction evaluates whether the gap is sufficient to create a pedagogically meaningful distinction—even for students at the same proficiency level.

To compute a DAS that reflects whether the difficulty difference was educationally meaningful, we combined the three annotators’ responses with a weighted sum that yields a value between $-1$ and $+1$:
\begin{align}
    \sum_{r=1}^{3} w_r a_r
\end{align}
where $a_r \in \{+1, -1\}$ denotes annotator~$r$’s pairwise judgment, and $w_r$ represents the category weight: $w_r=0.5$ for Case~1 and $w_r=1$ for Case~2.
If all three annotators agreed with the intended ordering and perceived distinguishable difficulty gap, the resulting score was $+1$, indicating perfectly aligned difficulty.
Conversely, if at least one annotator disagreed while marking Case~1 (minimal gap), the score approached $0$ (absolute value $\approx 0.1667$), suggesting negligible difficulty difference between the two items.

\paragraph{Complete Alignment Ratio.} This metric defined as the proportion of pairs for which all human evaluators unanimously agreed that the observed ordering matched the intended direction.
A higher CAR indicates a stronger ability to calibrate difficulty according to the specified feature constraints.

\paragraph{Validity.}
We assessed the validity of the generated questions including its answerability, answer matching and logical integrity considering that our targeting item format is MCFI (with the stem of ``Which statement is True based on the passage?''). We measured this on 3-point scale:
1 — The item is unanswerable (no answer or multiple answer), and the generated answer (statement with the factuality True) is not the correct answer.
2 — The item is answerable and the real answer and intended answer are matched. But the options are not meutualy independent (at least two options are the relationship of entailment or contradiction making the determination of the factuality of one option also determine the factuality of the counterpart without referencing the passage). This issue is not critical but degrades the item quality.
3 — Fully valid. the items are answerable and all options are well-constructed.
Validity was evaluated using GPT-5-mini by sampling eight times, and the average score was reported as follows~\citet{liu2023g}.

\begin{table*}[t]
\centering
\small
\begin{tabularx}{\textwidth}{c|ccc|X}
\toprule
Level & Vocabulary level & Passage length & Sentence length & Option Constraints (Factuality / Reasoning Complexity) \\ \midrule
1 & CEFR A--A2 & short & short & T / S-WM, F / S-WM, F / S-WM, F / S-WM \\
2 & CEFR A1--A2 & short & short & T / S-P, F / S-P, F / S-WM, F / S-WM \\
3 & CEFR A1--A2 & medium & short & T / S-P, F / S-P, F / S-WM, F / S-WM \\
4 & CEFR B1--B2 & medium & medium & T / S-P, F / S-P, F / S-P, F / S-P \\
5 & CEFR B1--B2 & medium & medium & T / S-I, F / S-I, F / S-P, F / S-P \\
6 & CEFR B1--B2 & medium & long & T / S-I, F / S-I, F / S-P, F / S-P \\
7 & CEFR C1--C2 & long & long & T / S-I, F / S-I, F / S-I, F / S-I \\
8 & CEFR C1--C2 & long & long & T / M-I, F / M-I, F / M-I, F / M-I \\ \bottomrule
\end{tabularx}
\caption{Eight-level difficulty-calibrated feature constraint sequence. (Abbreviations: T / True, F / False, S: Single-sentence evidence, M: Multi-sentence evidence, WM: Word Matching, P: Paraphrasing, I: Inference)}
\label{tab:difficulty_constraints}
\end{table*}

\paragraph{Coherence and Fluency.}  
To ensure readability and naturalness, we measured the coherence and fluency of generated passages using the UniEval framework, which are generally used automatic evaluation metrics that has high human alignment and largely used especially in text summarization task.

\section{\label{appendix:detail_calibrated_sequence} Constructing Difficulty-Calibrated Feature Constraint Sequence}

To construct the difficulty-calibrated feature constraint sequence utilized in our experiments, we first defined sixteen candidate constraint sets and generated corresponding items using \modelName\ based on fifteen source documents (disjoint from the test set).
We then computed the DAS for item pairs generated from constraint sets within a sliding window of size 5, ensuring that each item was compared with others of nearby theoretical difficulty, using GPT-5-mini as the judge with $N = 4$ in Equation~\ref{eq:da}.
Constraint pairs with a DAS below 0.4 were filtered out, and from the remaining candidates, we constructed the final sequence, which exhibited a monotonically increasing difficulty level.
Consequently, we obtained eight levels of difficulty-calibrated feature constraints that are both theoretically and empirically validated (see Table~\ref{tab:difficulty_constraints}).

\section{\label{appendix:das_std}Difficulty Alignment Score Statistics of LLM Judges}

Table~\ref{tab:das_statistics} presents the DAS statistics across 8 repeated inferences (4 for each of the forward and reversed comparisons).
Inherent variation across multiple samplings is reflected in the DAS, as higher variance leads the mean score closer to 0.
Beyond this, the STD values offer additional insights.
For instance, comparing Level-based Incremental Prompting$_{\mathrm{Qwen3\text{-}32B}}$ with MAFIG$_{\mathrm{Qwen3\text{-}32B}}$, the two methods show a substantial difference in DAS while exhibiting nearly identical STDs. 
This suggests that although the frequency of inconsistent judgments was similar, the Level-based Incremental model more frequently produced reversed difficulty judgments (i.e., scores approaching $-1$), which offset the positive scores.

\begin{table}[t]
\centering
\scriptsize
\begin{tabular}{llcc}
\toprule
\begin{tabular}[c]{@{}l@{}}Difficulty Control\\ Granularity\end{tabular} & Method  & Mean & STD \\
\midrule
\multirow{4}{*}{Level-based}
  & Direct$_{\mathrm{Qwen3\text{-}32B}}$      & 0.1037 & 0.5434 \\
  & Direct$_{\mathrm{GPT\text{-}5}}$          & 0.2949 & 0.6309 \\
  & Incremental$_{\mathrm{Qwen3\text{-}32B}}$ & 0.1804 & 0.4430 \\
  & Incremental$_{\mathrm{GPT\text{-}5}}$     & 0.2750 & 0.5706 \\
\midrule
\multirow{3}{*}{Feature-based}
  & Direct$_{\mathrm{Qwen3\text{-}32B}}$      & 0.2759 & 0.4941 \\
  & Direct$_{\mathrm{GPT\text{-}5}}$          & 0.4952 & 0.4502 \\
  & MAFIG$_{\mathrm{Qwen3\text{-}32B}}$       & 0.5229 & 0.4495 \\
\bottomrule
\end{tabular}
\caption{DAS statistics of LLM judges. The range of DAS is $[-1, 1]$.}
\label{tab:das_statistics}
\end{table}

\section{\label{appendix:human_evaluation_details}Human Evaluation Setup}

\begin{figure}[t]
    \centering
    \includegraphics[width=0.95\columnwidth]{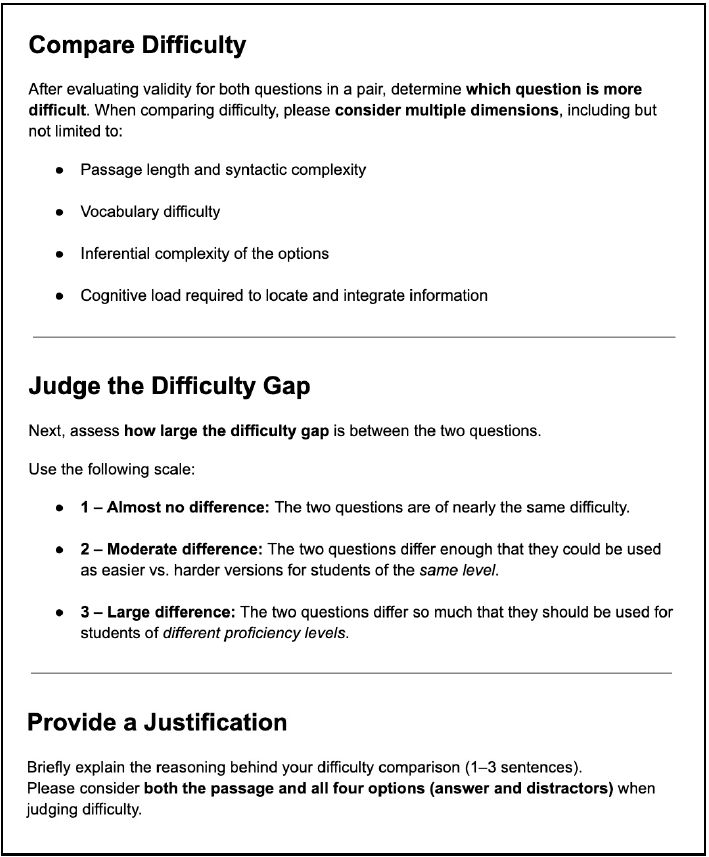}
    \caption{Screenshots of instructions provided to human evaluators.}
    \label{fig:instruction_sheet}
\end{figure}

We recruited three professional English instructors via Upwork\footnote{\url{https://www.upwork.com/}}, selecting candidates with proven expertise in IELTS instruction or RC item development.
The evaluation set comprised item pairs from adjacent difficulty levels (Level 1 to 8) sampled across six distinct source documents and three generation methods, resulting in 126 pairs per annotator.
Each evaluator was compensated with \$115 for the entire task and all participants provided informed consent that their anonymized evaluations may be released for research purposes.
Detailed annotation instructions are available in Figure~\ref{fig:instruction_sheet}.

\section{\label{appendix:feature-wise}Feature-wise Analysis}

\begin{figure*}[t]
\centering
\includegraphics[width=\textwidth]{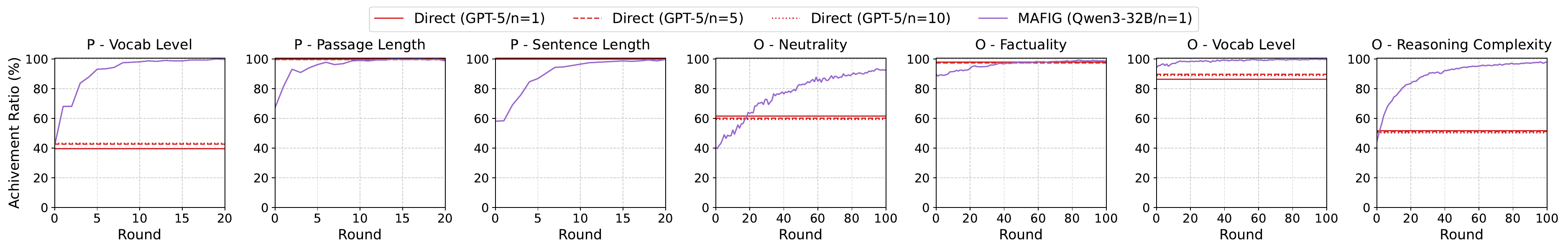}
\caption{\label{fig:feature_wise_analysis} Feature-wise ARs for GPT-5 (Direct Prompting) and Qwen3-32B-based \modelName, showing differential constraint satisfaction across feature types.}
\end{figure*}

\begin{figure*}[h!]
    \centering
    
    \begin{subfigure}{\linewidth}
        \centering
        \includegraphics[width=\linewidth]{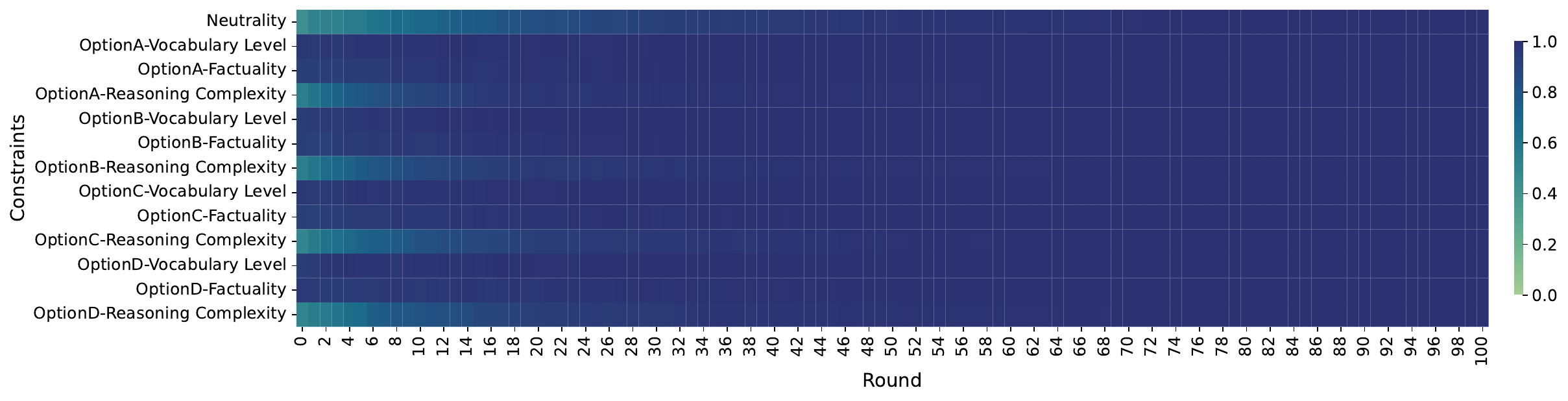}
        \caption{\label{subfig:a} Successful examples within 100 rounds.}
    \end{subfigure}
    
    \vspace{0.5em}
    
    \begin{subfigure}{\linewidth}
        \centering
        \includegraphics[width=\linewidth]{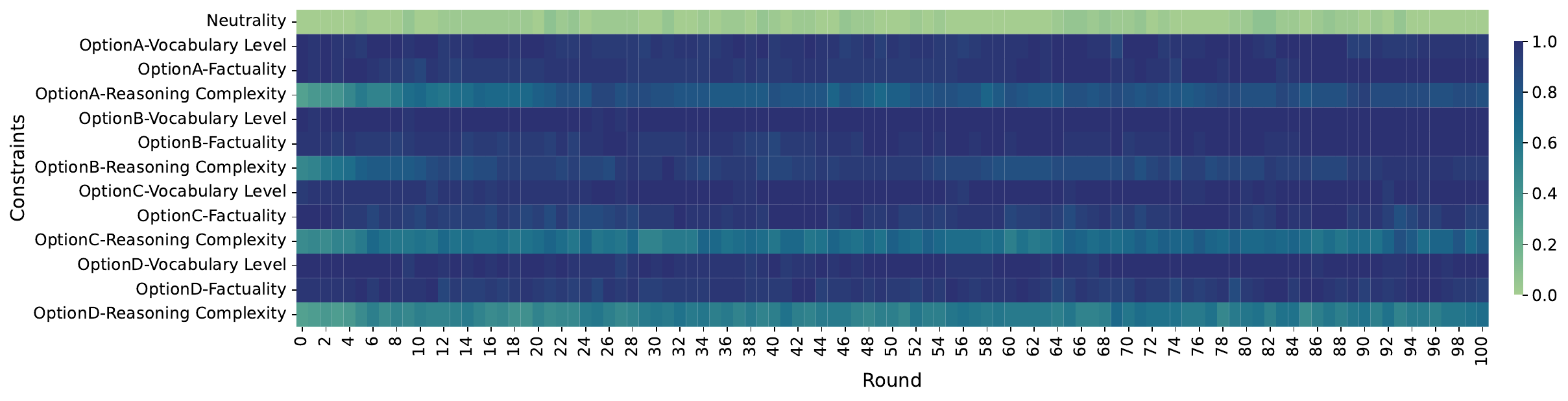}
        \caption{\label{subfig:b} Failed examples within 100 rounds.}
    \end{subfigure}
    
    \caption{Visualization of feature-wise satisfaction across rounds for examples that (a) succeed or (b) fail to achieve full constraint satisfaction within 100 rounds in the option generation stage. Each cell in the heatmap represents the proportion of examples satisfying a given constraint at a specific round, where values closer to 1 indicate higher success rates and values closer to 0 indicate failure across all examples.}
    \label{fig:feature_wise_constraint_satisfaction_proportion}
\end{figure*}

We compared the AR of GPT-5 under Direct Prompting and Qwen3-32B-based \modelName\ (with a single draft, $n=1$) in generating items that satisfy six feature constraints used for difficulty control.
For passage generation, we independently evaluated whether the generated passages satisfied the constraints on \textit{vocabulary level}, \textit{passage length}, and \textit{sentence length}.
For option generation, we assessed the satisfaction of \textit{neutrality} among options and, for each option, the constraints of \textit{factuality}, \textit{vocabulary level}, and \textit{reasoning complexity}.

As shown in Figure~\ref{fig:feature_wise_analysis}, GPT-5 achieved high satisfaction ratios for surface-level and factuality-based features—specifically, \textit{passage length}, \textit{sentence length}, and \textit{factuality}.
However, its SR sharply declined for the \textit{vocabulary level} constraint, which requires alignment with an external lexical standard.
Similarly, GPT-5 exhibited limited performance in satisfying deeper cognitive constraints, such as \textit{neutrality} among options and \textit{reasoning complexity}, both of which require analyzing cross-option relationships and inferential reasoning beyond surface features.

These findings indicate that while with a reasoning-optimized model such as GPT-5 is sufficient to generate items aligned with surface-level or factual constraints, revision through \modelName\ is essential for constraints that depend on external standards or demand more intricate cognitive control.
Notably, GPT-5’s overall ARs were generally higher than those of Qwen3-32B before revision, suggesting that employing GPT-5 as an agent backbone within \modelName\ could enable faster convergence toward fully constraint-satisfying items.

\section{\label{appendix:case_study}Case Study}

\begin{table}[t]
\centering
\scriptsize
\begin{tabular}{p{7.2cm}}
\toprule
\textbf{Level 4} \\
\midrule
Moral decisions often involve difficult choices.
Sometimes, an action may cause both good and bad results.
People must decide whether it is right to allow the bad effect if the main goal is good.
This idea is called the principle of double effect.
It says that if an action directly causes good and indirectly causes harm, it may be acceptable.
But the harm must not be the main goal.
The person must know the harm will happen, but still choose the action for the good reason.
For example, a doctor might give a treatment that helps a patient but also causes pain.
The pain is not the main goal, but it is known to happen.
Some people argue that certain actions, like war, should never be allowed if they could destroy the human race.
They believe no reason is big enough to justify such destruction.
Others say that avoiding war at all costs could also be wrong.
For example, giving in to an enemy might lead to long-term suffering.
The key is to weigh the good and bad effects carefully.
People must act based on what is right now, not just what might happen later.
Moral choices are hard, but they must be made with clear thinking and responsibility. \\ \midrule
According to the passage, which of the following statements is true? \\
A. The principle of double effect allows an action if the harm is directly caused and the good is indirect. \\
B. People must be aware of the harm but still act for a good reason. \\
C. People should only consider future consequences when making moral decisions. \\
D. Moral decisions should be based only on what might happen in the future, not on what is right now. \\
\bottomrule
\end{tabular}
\caption{\label{tab:fail_example} An MCFI item that fails to satisfy the feature constraints for Level 4. After 100 rounds of revision, the item still violates the neutrality and reasoning complexity constraints for options A and C.}
\end{table}

Through the preceding experiments, we observed that passage generation was able to achieve constraint satisfaction in nearly all examples within 20 rounds (or within 5 rounds when the draft size was set to 5), whereas option generation occasionally failed to reach full satisfaction even after 100 rounds. 
In this section, we analyze what poses the greatest obstacles to constraint satisfaction in the option generation stage.

Figure~\ref{fig:feature_wise_constraint_satisfaction_proportion} illustrates which feature constraints posed the greatest obstacles to full satisfaction.
Results from both successful and failed cases indicate that the model consistently struggled to satisfy the neutrality and the reasoning complexity constraints.
Notably, in cases where full satisfaction was not achieved even after 100 rounds, neutrality was found to be the constraint most persistently violated.

A representative case of such neutrality violations is presented in Table~\ref{tab:fail_example}, where the target item was generated at Level 4.
The passage associated with this item addresses the topic of ``moral decision-making'' and is written with a high degree of logical cohesion. 
Because the sentences within the passage are mutually interdependent, generating contradicted statements that maintain a neutral relationship with one another proves particularly challenging.
This suggests that the topic itself gives rise to a passage more suited to higher-order comprehension tasks — such as Summarization or Main Idea Identification formats — than to the Factual Information format.
In our experiments, the topic of each item was controlled through the source text, but was not treated as a variable requiring explicit control for difficulty calibration. 
In practical applications, incorporating passage topic as an additional factor in the item generation process would likely mitigate constraint satisfaction failures attributable to such misalignment between passage topic and item type.

\section{\label{appendix:human_justification_analysis}Qualitative Analysis of Difficulty Factors}

\begin{table*}[h!]
\centering
\scriptsize
\begin{tabular}{p{3cm}|p{1cm}|p{10.5cm}}
\toprule
Label                                     & Frequency & Examples  \\ \midrule
Answer verification difficulty            & 57.85\%                         & demands paraphrasing skills, requires inferential understanding, answer directly stated, inference across multiple sentences, answer options use the identical wording as in the passage, direct recall, evidence from a single statement          \\ \midrule
Passage vocabulary difficulty             & 50.52\%                         & more advanced vocabulary, familiar vocabulary, complex vocabulary, difficult words, moderate vocabulary                                                                                                                                            \\ \midrule
Passage syntactic / structural complexity & 45.81\%                         & longer sentences, more complex sentence structures, easier to read, compound and complex sentence structures, complex syntax, denser syntax, higher syntactic complexity, complex kind of sentences, simple sentences                              \\ \midrule
Passage length / word count               & 45.29\%                         & length of the both questions almost same, lengthier, lengthier passage, greater word count, significantly longer, long passage                                                                                                                      \\ \midrule
Figurative language / tone / metaphors    & 3.66\%                          & requires readers to make emotional connections, interpretation of the emotional tone from the dialogue, literal meaning, literal comprehension, metaphor (nightingale) requiring inferential grasp, vocabualry laden with metaphorical connotation \\ \midrule
Option vocabulary difficulty              & 3.14\%                          & options use advanced vocabulary, options use very advanced vocabulary, the options in Q1 contain more advanced vocabulary                                                                                                                          \\ \midrule
Option length / word count                & 2.88\%                          & wordy statements, wordier options, short answer option, wordier options (A, B, and D), lengthier options \\

\bottomrule

\end{tabular}
\caption{\label{tab:expert_observations} Distribution of perceived difficulty factors derived from qualitative expert feedback.}
\end{table*}

To explore whether feature-based difficulty control effectively translates into actual difficulty calibration, we conducted a qualitative analysis based on the justifications provided by three domain experts during the pairwise evaluation. 
Experts were asked to specify the underlying factors that influenced their perception of difficulty for each item pair.

\paragraph{Clustering of Perceived Difficulty Factors.}
We employed a three-stage prompting pipeline using GPT-5 to systematically cluster the expert justifications. 
First, we extracted general difficulty-related keywords and phrases from each justification. 
Second, the model was prompted to generate representative cluster labels by synthesizing these extracted factors. 
After manual refinement, we performed a final classification step where each initial factor was mapped to its most appropriate cluster.

As summarized in Table~\ref{tab:expert_observations}, the \textit{answer verification difficulty} cluster—which encompasses factors related to mapping between the passage and options, as well as reasoning complexity—was the most prevalent, accounting for 57.85\% of expert mentions.
This aligns closely with the reasoning complexity feature we explicitly controlled in our experiments. While this dimension is particularly challenging to satisfy via single-pass prompting (as discussed in Section~\ref{appendix:feature-wise}), our results confirm that it is the most critical factor used by experts to calibrate or assess item difficulty.
Conversely, factors not directly targeted by our framework, such as figurative language, tone, and option length, were also noted (approximately 6\%), suggesting that these unmonitored features may introduce unintended variations in difficulty.

\begin{figure}[t]
    \centering
    \includegraphics[width=\columnwidth]{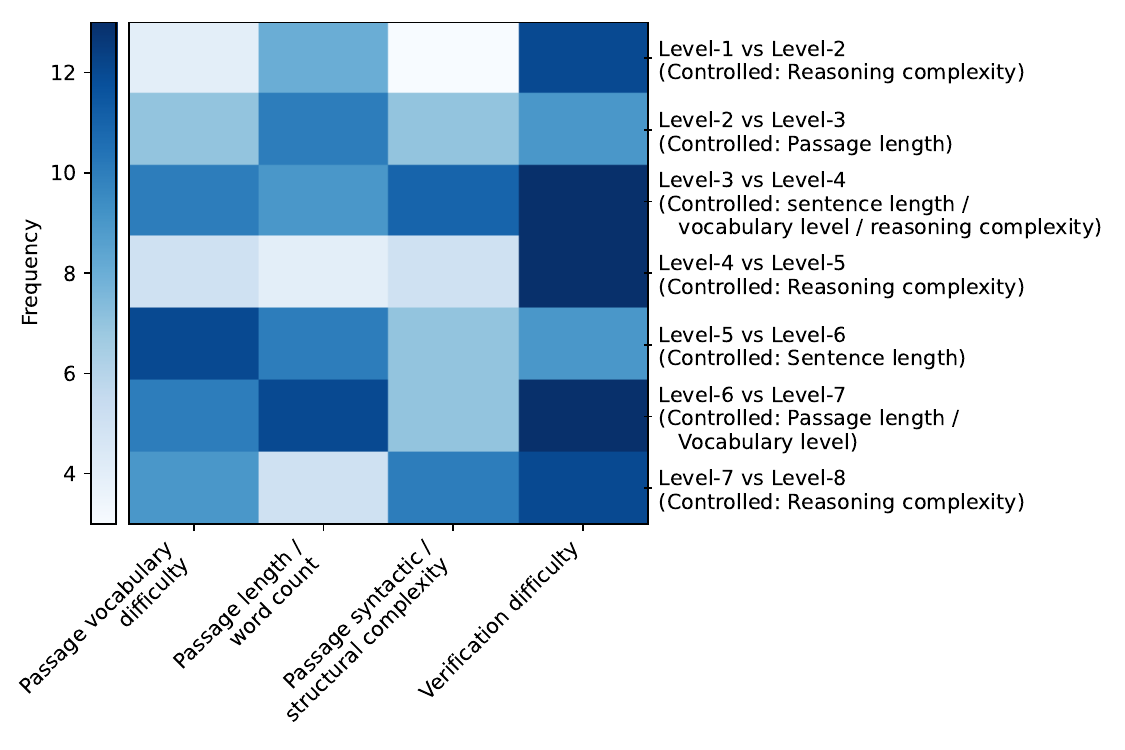}
    \caption{Expert-perceived difficulty factors for each level pair.}
    \label{fig:controlled_vs_perceived}
\end{figure}

\paragraph{Alignment between Controlled Features and Expert Perception.}

Figure~\ref{fig:controlled_vs_perceived} visualizes the relationship between the intended difficulty calibration in our eight-level feature constraint sequence and the factors actually perceived by experts. 
In our experimental setup, adjacent difficulty levels were differentiated by intentionally increasing the cognitive complexity of specific features.

The visualization reveals strong alignment in pairs where reasoning complexity was the intended differentiator (e.g., Level 1–2, Level 3–4, Level 4–5, and Level 7–8); in these cases, experts frequently cited factors belonging to the \textit{answer verification difficulty} cluster. 
Interestingly, this cluster was also prominent in the Level 6–7 pair, where passage length was the primary variable; we hypothesize that increased passage length made locating evidence for options more cognitively demanding, thereby indirectly affecting verification difficulty. 
Furthermore, intended increases in vocabulary level (Level 3–4, Level 5–7) and adjustments in sentence and passage length were accurately identified by experts as primary drivers of difficulty disparity.
Consequently, \modelName\ can effectively support item writers to modulate the difficulty of RC items by precisely manipulating specific features in accordance with their pedagogical intentions.

However, we observed instances where unintended features were cited as difficulty factors. 
This likely stems from inherent correlations between linguistic features—for example, increasing sentence length often leads to more complex syntactic structures and the inclusion of more sophisticated vocabulary. 
While our framework manages features independently, these results suggest that modeling the interdependencies between difficulty factors is a promising avenue for future improvement of the system.

\section{\label{appendix:prompt_templates}Prompt Templates}

Figures~\ref{fig:drafter},~\ref{fig:planner},~\ref{fig:editor},~\ref{fig:reworder}, and~\ref{fig:refiner} present the prompt templates used during the passage generation stage. The templates used during the option generation stage and for the LLM-based evaluation can be found in our GitHub repository: \url{https://github.com/SeonjeongHwang/mafig}.

%% Drafter
\begin{figure*}[ht]
\scriptsize
\begin{tabular}{p{0.95\textwidth}}
\toprule
\textbf{Drafter} \\ \midrule
You are an item writer for reading comprehension tests. \\
Your task is to write a reading passage. \\
The passage must be based entirely on factual information from a given source text and meet three constraints: passage length, sentence length, and vocabulary level. \\
The passage is not a summary of the source text, but a reading material suitable for reading comprehension tests. \\ \vspace{0.05em}

\textbf{\#\#\# Constraint Definitions:} \\
1. Passage Length -- Total number of sentences in the passage: \\
\quad - short: 5--10 sentences \\
\quad - medium: 11--20 sentences \\
\quad - long: 21--30 sentences \\ \vspace{0.05em}

2. Sentence Length -- Average number of words per sentence: \\
\quad - short: 10 words or fewer \\
\quad - medium: more than 10 words and less than or equal to 15 words \\
\quad - long: more than 15 words and less than or equal to 20 words \\ \vspace{0.05em}

3. Vocabulary Level -- CEFR level of all words used in the passage: \\
\quad - A: A1--A2 words only \\
\quad - B: B1--B2 or simpler \\
\quad - C: C1--C2 or simpler \\
\quad - The passage must include at least one word from the specified level, and must not use any words above that level. \\ \vspace{0.05em}

\textbf{\#\#\# Step-by-Step Guide:} \\
\textbf{Step 1: Compression of Content} \\
Because the source text is lengthy and contains extensive information, you should not attempt to retell everything or summarize all of it. Instead, focus on one specific aspect of the text---for example, an event, a scene, or a key piece of information. \\ \vspace{0.05em}

\textbf{Step 2: Adapt Writing Type and Style Freely} \\
Check the writing type, style, and sentence length of the source text. \\
You are not required to keep the same type as the source (narrative, expository, descriptive, persuasive, etc.). \\
You may also change the style (dialogue-heavy, descriptive, instructional, advertisement-style, etc.) if it helps meet constraints. \\ \vspace{0.05em}

\textbf{Step 3: Passage Planning} \\
Explain how you will structure the passage: how it will begin, develop, and end. Describe the logical flow. \\ \vspace{0.05em}

\textbf{Step 4: Length Control} \\
Fix the number of sentences (within the range), and assign content evenly. \\ \vspace{0.05em}

\textbf{Step 5: Sentence \& Vocabulary Control} \\
Adjust sentence length to meet word-count rules (add modifiers/examples if short, condense if long). Use only vocabulary at or below the CEFR target level, and include at least one word from the required level. \\ \vspace{0.05em}

\textbf{Step 6: Fidelity Check} \\
Confirm all statements are consistent with the source text. \\
Exclude, merge, or rephrase details that don't fit the target form. \\ \vspace{0.05em}

\textbf{\#\#\# Output Format:} \\
\texttt{\{"passage": "..."\}} \\ \vspace{0.05em}

\textbf{\#\#\# Input:} \\
Source Text: \\
\colorbox{gray!30}{\{source\_ document\}} \\ \vspace{0.05em}

Constraints: \\
- Passage Length: \colorbox{gray!30}{\{passage\_length\}} \\
- Sentence Length: \colorbox{gray!30}{\{sentence\_length\}} \\
- Vocab Level: \colorbox{gray!30}{\{vocab\_level\}} \\ \vspace{0.05em}

\textbf{\#\#\# Output:} \\
First, explain your thought process in detail. Then, provide the final output in the specified JSON format. \\
Let's think step by step. \\
\bottomrule
\end{tabular}
\caption{Prompt template for the \textbf{Drafter} agent used in the passage generation stage.
\colorbox{gray!30}{\{placeholder\}} indicates a slot to be filled with the corresponding value.}
\label{fig:drafter}
\end{figure*}

%% Planner
\begin{figure*}[ht]
\scriptsize
\begin{tabular}{p{0.95\textwidth}}
\toprule
\textbf{Planner} \\ \midrule
You are an item planner for reading comprehension tests. \\
Your task is to decide which action to take in order to revise the passage based on: \\
\quad - A source text that is the original factual content that the passage must be based on \\
\quad - messages you sent to other agents in previous trials \\
\quad - A current version of the passage that you are editing \\
\quad - An error report that shows which constraints the current passage violates \\ \vspace{0.05em}

Your task is instructions for agents to ensure that the passage satisfies all given constraints. \\
Decide which action to take. You have two options: \\
\quad 1) \textbf{Call\_Editor} \\
\quad\quad - The Editor is responsible for performing holistic revision of the passage, with the exception of vocabulary-level control. \\
\quad\quad - Send a message containing detailed guidance on the current issues, based on the given evaluation report and previous messages. \\
\quad\quad - Do not direct the Editor to apply vocabulary of a specific CEFR level, as the Editor does not have access to a CEFR dictionary. \\
\quad\quad - If you fail \colorbox{gray!30}{\{threshold\}} times in a row with the same approach, you need to try a more creative approach. \\
\quad 2) \textbf{Call\_Reworder} \\
\quad\quad - The Reworder performs lexicon-only substitutions to satisfy CEFR-based vocabulary constraints. \\
\quad\quad - Send a message identifying which words are problematic. \\
\quad\quad - Since the planner (you) do not know the exact vocabulary level of these words, you must not suggest any synonyms. \\ \vspace{0.05em}

When you call the agents, you must send them detailed instructions. \\
Review the messages from prior trials and the error report for the current passage, and give the agents clear guidance to avoid repeating the same mistakes. \\ \vspace{0.05em}

\textbf{\#\#\# Constraint Definitions:} \\
1. Passage Length -- Total number of sentences in the passage: \\
\quad - short: 5--10 sentences \\
\quad - medium: 11--20 sentences \\
\quad - long: 21--30 sentences \\ \vspace{0.05em}

2. Sentence Length -- Average number of words per sentence: \\
\quad - short: 10 words or fewer \\
\quad - medium: more than 10 words and less than or equal to 15 words \\
\quad - long: more than 15 words and less than or equal to 20 words \\ \vspace{0.05em}

3. Vocabulary Level -- CEFR level of all words used in the passage: \\
\quad - A: A1--A2 words only \\
\quad - B: B1--B2 or simpler \\
\quad - C: C1--C2 or simpler \\
\quad - The passage must include at least one word from the specified level, and must not use any words above that level. \\ \vspace{0.05em}

\textbf{\#\#\# Output Format:} \\
When you decide to call the Editor, return the following JSON: \\
\texttt{\{"action": "Call\_Editor", "message": "An instruction for the Editor except vocabulary issues."\}} \\ \vspace{0.05em}

When you decide to call the Reworder, return the following JSON: \\
\texttt{\{"action": "Call\_Reworder", "message": "An instruction for the Reworder"\}} \\ \vspace{0.05em}

\textbf{\#\#\# Input:} \\
Source Text: \\
\colorbox{gray!30}{\{source\_document\}} \\ \vspace{0.05em}

Past Messages You Sent to Agents: \\
\colorbox{gray!30}{\{revision\_memory\}} \\ \vspace{0.05em}

Current Passage: \\
\colorbox{gray!30}{\{current\_state\}} \\ \vspace{0.05em}

Constraints: \\
\quad - Passage Length: \colorbox{gray!30}{\{passage\_length\}} \\
\quad - Sentence Length: \colorbox{gray!30}{\{sentence\_length\}} \\
\quad - Vocabulary Level: \colorbox{gray!30}{\{vocab\_level\}} \\ \vspace{0.05em}

Error Report: \\
\colorbox{gray!30}{\{error\_report\}} \\ \vspace{0.05em}

\textbf{\#\#\# Output:} \\
First, explain your thought process in detail. Then, provide the final output in the specified JSON format. \\
Let's think step by step. \\
\bottomrule
\end{tabular}
\caption{Prompt template for the \textbf{Planner} agent used in the passage generation stage.
\colorbox{gray!30}{\{placeholder\}} indicates a slot to be filled with the corresponding value.}
\label{fig:planner}
\end{figure*}

%% Editor
\begin{figure*}[ht]
\scriptsize
\begin{tabular}{p{0.95\textwidth}}
\toprule
\textbf{Editor} \\ \midrule
You are an item editor for reading comprehension tests who should change the passage based on the item planner's supervision. \\
You will be provided with the following information: \\
\quad - A source text that is the original factual content that the passage must be based on \\
\quad - A current version of the passage that you are revising \\
\quad - A set of constraints that the passage must satisfy \\
\quad - An instruction from the item planner \\ \vspace{0.05em}

\textbf{\#\#\# Constraint Definitions:} \\
1. Passage Length -- Total number of sentences in the passage: \\
\quad - short: 5--10 sentences \\
\quad - medium: 11--20 sentences \\
\quad - long: 21--30 sentences \\ \vspace{0.05em}

2. Sentence Length -- Average number of words per sentence: \\
\quad - short: 10 words or fewer \\
\quad - medium: more than 10 words and less than or equal to 15 words \\
\quad - long: more than 15 words and less than or equal to 20 words \\ \vspace{0.05em}

\textbf{\#\#\# Output Format:} \\
Return the passage segmented by sentences. \\
\texttt{\{"sentence 1": "<sentence 1>", "sentence 2": "<sentence 2>", ...\}} \\ \vspace{0.05em}

\textbf{\#\#\# Input:} \\
Source Text: \\
\colorbox{gray!30}{\{source\_document\}} \\ \vspace{0.05em}

Current Passage: \\
\colorbox{gray!30}{\{current\_state\}} \\ \vspace{0.05em}

Constraints: \\
\quad - Passage Length: \colorbox{gray!30}{\{passage\_length\}} \\
\quad - Sentence Length: \colorbox{gray!30}{\{sentence\_length\}} \\ \vspace{0.05em}

Planner's Instruction: \\
\colorbox{gray!30}{\{planner\_instruction\}} \\ \vspace{0.05em}

\textbf{\#\#\# Output:} \\
First, explain your thought process in detail. Then, provide the final output in the specified JSON format. \\
Let's think step by step. \\
\bottomrule
\end{tabular}
\caption{Prompt template for the \textbf{Editor} agent used in the passage generation stage.
\colorbox{gray!30}{\{placeholder\}} indicates a slot to be filled with the corresponding value.}
\label{fig:editor}
\end{figure*}

%% Reworder
\begin{figure*}[ht]
\scriptsize
\begin{tabular}{p{0.95\textwidth}}
\toprule
\textbf{Reworder -- Suggestion Phase} \\ \midrule
You are an item reworder for reading comprehension tests who should revise a specific item based on the item writing planner's supervision. \\
Your task is to identify words that violate the target CEFR level, and suggest alternative words that would satisfy the level while preserving the original meaning. \\
You will receive: \\
\quad - The source text, so you can ensure that any replacement words preserve the original meaning \\
\quad - A CEFR target level \\
\quad - A list of sentences from a passage \\
\quad - An instruction from the item planner \\ \vspace{0.05em}

\textbf{\#\#\# Constraint Definitions:} \\
\quad - A: A1--A2 words only \\
\quad - B: B1--B2 or simpler \\
\quad - C: C1--C2 or simpler \\
The passage must include at least one word from the specified level, and must not use any words above that level. \\ \vspace{0.05em}

\textbf{\#\#\# Output Format:} \\
For each sentence, the problematic words and their acceptable replacements, including both: \\
\quad - replacements for words that are too difficult, and \\
\quad - upgrade suggestions for words that are too easy when necessary. \\
\texttt{\{"1": \{"appointed": ["named", "put in"], "resigned": ["left", "quit"]\}, "2": \{"good": ["excellent", "strong"]\}\}} \\ \vspace{0.05em}

\textbf{\#\#\# Input:} \\
Source Text: \\
\colorbox{gray!30}{\{context\}} \\ \vspace{0.05em}

Passage: \\
\colorbox{gray!30}{\{current\_state\}} \\ \vspace{0.05em}

Target CEFR Level: \colorbox{gray!30}{\{vocab\_level\}} \\ \vspace{0.05em}

Planner's Instruction: \\
\colorbox{gray!30}{\{planner\_instruction\}} \\ \vspace{0.05em}

\textbf{\#\#\# Output:} \\
First, explain your thought process in detail. Then, provide the final output in the specified JSON format. \\
Let's think step by step. \\
\bottomrule
\toprule
\textbf{Reworder -- Replacement Phase} \\ \midrule
You are an item reworder for reading comprehension tests. \\
Your task is to revise a passage so that all vocabulary complies with the specified CEFR difficulty level. \\
You will receive: \\
\quad - A list of sentences from a passage \\
\quad - For each sentence: \\
\quad\quad - A dictionary of problematic words (i.e., above or below the allowed CEFR level) \\
\quad\quad - For each word, a list of replacement candidates (may be empty) \\
\quad - The CEFR \texttt{target\_level} the passage must conform to (\texttt{A}, \texttt{B}, or \texttt{C}) \\ \vspace{0.05em}

\textbf{\#\#\# Rewording Rules} \\
\quad - All words must be at or below the given CEFR level. \\
\quad - The passage must include at least one word from the exact target level. \\
\quad - Only use replacement words from the provided list. \\
\quad - Do not generate any new words or phrases not in the replacement list. \\ \vspace{0.05em}

\textbf{\#\#\# Output Format:} \\
\texttt{\{"updated": ["revised version of sentence 1", "revised version of sentence 2", ...],} \\
\texttt{"message": "The word 'engage' in sentence (3) cannot be reworded. ..."\}} \\ \vspace{0.05em}

\textbf{\#\#\# Input:} \\
Passage: \\
\colorbox{gray!30}{\{current\_state\}} \\ \vspace{0.05em}

Target CEFR Level: \colorbox{gray!30}{\{vocab\_level\}} \\ \vspace{0.05em}

Replacement Candidates: \\
\colorbox{gray!30}{\{alternative\_list\}} \\ \vspace{0.05em}

\textbf{\#\#\# Output:} \\
First, explain your thought process in detail. Then, provide the final output in the specified JSON format. \\
Let's think step by step. \\
\bottomrule
\end{tabular}
\caption{Prompt template for the \textbf{Reworder} agent used in the passage generation stage.
\colorbox{gray!30}{\{placeholder\}} indicates a slot to be filled with the corresponding value.}
\label{fig:reworder}
\end{figure*}

%% Refiner
\begin{figure*}[ht]
\scriptsize
\begin{tabular}{p{0.95\textwidth}}
\toprule
\textbf{Refiner} \\ \midrule
You are a refiner. \\
Your task is to improve the coherence and fluency of a given passage. \\ \vspace{0.05em}

\textbf{Strict Constraints:} \\
\quad - You must preserve the original meaning and vocabulary exactly as given. \\
\quad - Do not add, remove, or replace any words. \\
\quad - You may only make minimal grammatical corrections and adjust sentence connections to improve readability. \\
\quad - Avoid stylistic rewriting, paraphrasing, or rewording. Only refine. \\ \vspace{0.05em}

\textbf{\#\#\# Output Format:} \\
\texttt{\{"passage": "..."\}} \\ \vspace{0.05em}

\textbf{\#\#\# Input:} \\
Passage: \\
\colorbox{gray!30}{\{passage\}} \\ \vspace{0.05em}

\textbf{\#\#\# Output:} \\
First, explain your thought process in detail. Then, provide the final output in the specified JSON format. \\
Let's think step by step. \\
\bottomrule
\end{tabular}
\caption{Prompt template for the \textbf{Refiner} agent used in the passage generation stage.
\colorbox{gray!30}{\{placeholder\}} indicates a slot to be filled with the corresponding value.}
\label{fig:refiner}
\end{figure*}

%% Evaluators

\end{document}